%% file: VLT-main.tex
\documentclass{IEEEtran}
\usepackage{multirow}
\usepackage{array}
\usepackage[utf8]{inputenc}
\usepackage{newunicodechar}
\newunicodechar{，}{,}

\usepackage{amsmath,amssymb,amsfonts}

\usepackage{subcaption}
\usepackage{cite}
\usepackage[ruled,linesnumbered]{algorithm2e}

\usepackage{color}
\usepackage{graphicx}
\usepackage{textcomp}
\usepackage{booktabs}
\usepackage{physics}
\usepackage{bbding}
\usepackage[section]{placeins}
\usepackage{tabularx}
\usepackage{float}
\usepackage{balance}
\usepackage{stfloats}
\usepackage{orcidlink}
\usepackage{hyperref}
\def\BibTeX{{\rm B\kern-.05em{\sc i\kern-.025em b}\kern-.08em
	T\kern-.1667em\lower.7ex\hbox{E}\kern-.125emX}}

\begin{document}

\title{
    \vspace{-1.5cm}
    \textnormal{
        \begin{minipage}{\textwidth}
            \centering
            \small
            This work has been submitted to IEEE for possible publication. Copyright may be transferred without notice, after which this version may no longer be accessible.
        \end{minipage}
    } \\
    \vspace{0.8cm}
    VLT: A Vision-Language-Time Series Multimodal Foundation Model for Industrial Intelligence
}

\author{
{
Haiteng Wang~\orcidlink{0000-0002-7316-3607},~\IEEEmembership{Member,~IEEE,}
Jingheng Yan~\orcidlink{0009-0002-8844-402X},~\IEEEmembership{Student Member,~IEEE,}
Xiaokang Wang~\orcidlink{0000-0002-0981-6204},~\IEEEmembership{Member,~IEEE,}
and Lei Ren~\orcidlink{0000-0001-6346-6930},~\IEEEmembership{Senior Member,~IEEE,}

}

\thanks{
    {	
The research is supported by The NSFC (National Science Foundation of China) project No.62225302, 623B2014, 62173023. (Haiteng Wang and Jingheng Yan contributed equally to this work.)(Corresponding authors: Lei Ren.)

Haiteng Wang, and Jingheng Yan are with the School of Automation Science and Electrical Engineering, Beihang University, Beijing 100191, China. (email: wanghaiteng@buaa.edu.cn, yjh967@buaa.edu.cn).

Lei Ren is with the School of Automation Science and Electrical Engineering, Beihang University, Beijing 100191, China, also with the Hangzhou International Innovation Institute, Beihang University, Hangzhou 311115, China, and also with the State Key Laboratory of Intelligent Manufacturing System Technology, Beijing 100854, China. (e-mail: renlei@buaa.edu.cn)

Xiaokang Wang is a Professor with the School of Computer Science and Artificial Intelligence, Zhengzhou University, Zhengzhou 450001, China. (e-mail: kxwang@zzu.edu.cn)
    }
}
}

\maketitle

\begin{abstract}
Industrial time-series serve as the foundation for Prognostics and Health Management (PHM) to ensure the reliability and safety of industrial equipment such as aero-engines. However, existing approaches are typically limited to single-modality modeling, which restricts their generalization in complex scenarios. Although recent advances in large language models (LLMs) provide new opportunities for multimodal learning, bridging continuous time-series signals and discrete textual semantics remains an open challenge. To this end, we propose VLT, a multimodal foundation model that jointly models time-series, frequency  visual representations, and textual knowledge. A key insight is to utilize frequency-spectrum as a visual bridge to connect continuous temporal signals with discrete semantics. Specifically, a Time-aware Mixture-of-Experts (Time-MoE) is designed to capture heterogeneous temporal dynamics, while a Frequency–Text Augmented Learner enables joint modeling of spectral and semantic features within a shared representation space. Furthermore, a time-centric gradient alignment mechanism is introduced to mitigate cross-modal optimization conflicts via gradient normalization and reliability-aware dynamic reweighting. Extensive experiments on multiple industrial datasets demonstrate that VLT outperforms state-of-the-art methods, achieving superior robustness and generalization under few-shot, noisy, and incomplete-modality settings.
\end{abstract} 


\section{Introduction}

Prognostics and Health Management (PHM) plays a crucial role in
improving the reliability and sustainability of industrial systems,
including aero-engines and chemical process systems. Data-driven PHM
commonly relies on multivariate sensor measurements to characterize
system health conditions and degradation processes. Under changing
operating conditions, cross-domain sensor alignment can reduce
distribution discrepancies in multivariate time-series representations
\cite{wang2024sea}. These measurements support critical tasks such as
remaining useful life (RUL) prediction \cite{zhu2026rul} and
time-series anomaly detection \cite{ho2025graph}, thereby providing
valuable information for condition-based maintenance and operational
decision-making.

\begin{figure}[t]
    \centering
    \setlength{\belowcaptionskip}{-0.8cm}  % 只对这张图生效
    \includegraphics[width=1.05\linewidth]{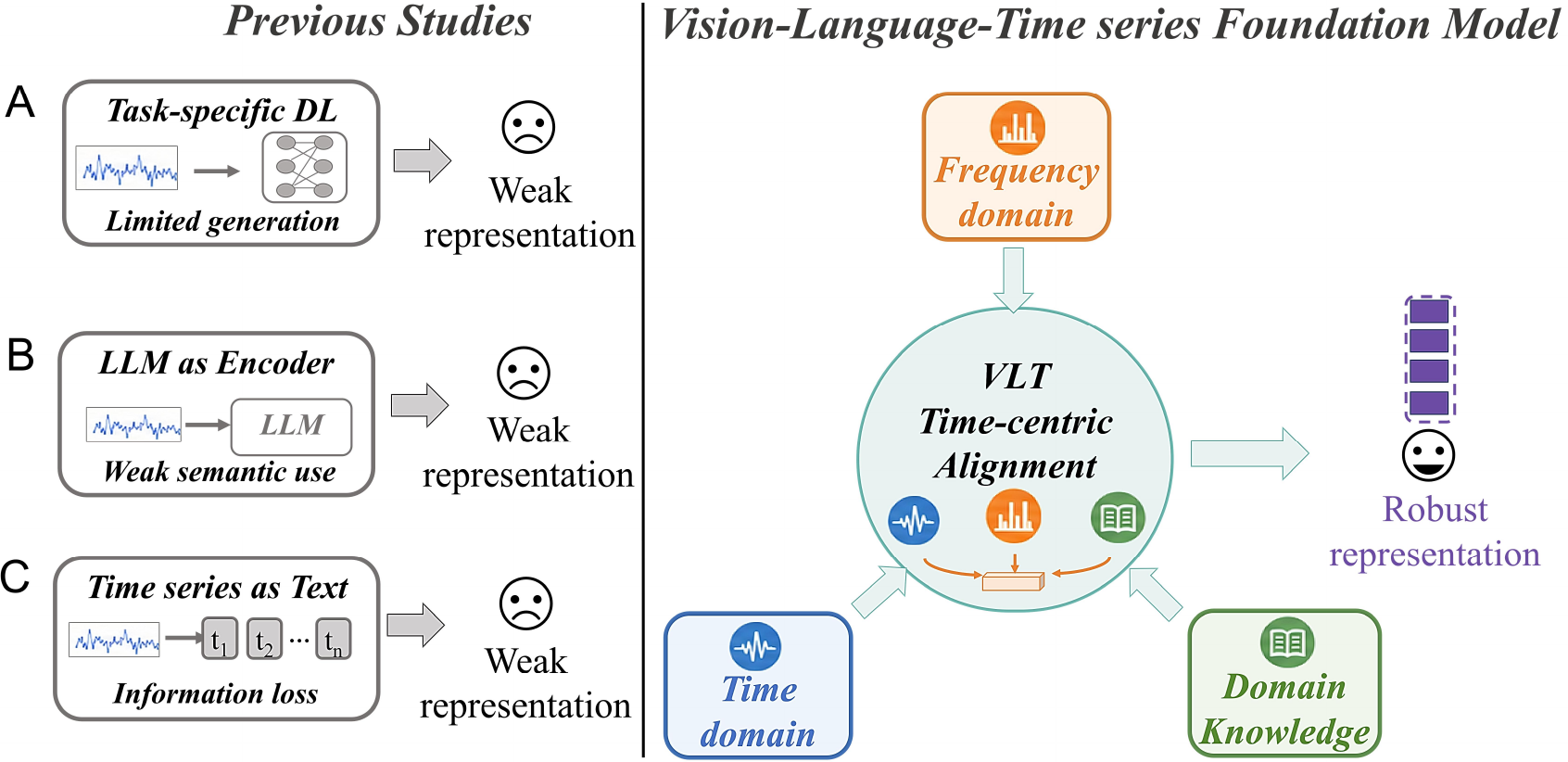}
    \caption{Comparison of existing PHM paradigms and the proposed VLT framework. Existing methods may have limited generalization, weak semantic use, or information loss. VLT improves PHM representation learning by aligning time-domain signals, frequency-domain features, and domain knowledge.}
    \label{fig:vlt_comparison}
\end{figure}

In recent years, PHM research has gradually expanded from
task-specific deep learning methods to Large Language Model
(LLM)-based paradigms. Existing approaches can be broadly classified
into three categories: task-specific deep learning methods, LLM-based
feature encoders, and LLM-driven time-series-to-language methods.

The first category consists of task-specific deep learning methods,
which employ convolutional, recurrent, attention-based, or generative
architectures to extract degradation features and model temporal
dependencies. More broadly, self-supervised time-series representation
learning has been systematically organized into generative,
contrastive, and adversarial paradigms \cite{zhang2024ssl}.
Diffusion-based models have also been explored for learning complex
temporal distributions \cite{ren2024mts,wang2025meta}. Although these
methods can achieve strong performance on in-domain data, their
effectiveness may deteriorate under distribution shifts or previously
unseen operating conditions \cite{wang2024sea,zhu2026rul}.

Furthermore, pretrained LLMs
\cite{hurst2024gpt,yang2025qwen3,guo2025deepseek} have emerged as
promising tools for time-series analysis. The second category uses
LLMs as feature encoders to process time-series data
\cite{yu2023harnessing,wang2025collm,tao2025llm}. Representative
methods include CoLLM \cite{wang2025collm} and One Fits All
\cite{zhou2023one}. These methods generally project numerical signals
into latent representations that can be processed by pretrained
language models. However, directly mapping numerical signals into the
LLM representation space may not fully exploit the linguistic
knowledge and semantic reasoning capabilities of LLMs.

The third category treats time series as natural language through
prompting, as exemplified by PromptCast \cite{xue2023promptcast} and
ChatTime \cite{wang2025chattime}, or incorporates auxiliary textual
information associated with continuous signals
\cite{jin2024time,lan2025gem,chen2025domain,wang2025diagllm}.
Although these methods allow LLMs to process time-series information
in a language-compatible form, converting continuous numerical
sequences into discrete textual tokens may introduce representation
mismatch and cause the loss of fine-grained temporal information.

Figure~\ref{fig:vlt_comparison}  compares existing PHM paradigms with the proposed VLT framework. Unlike previous methods that mainly rely on single-modality modeling or time-series-to-language conversion, VLT jointly exploits temporal signals, frequency-domain visual representations, and textual knowledge for more comprehensive PHM representation learning. Despite progress of LLMs in general time-series, existing PHM methods are typically limited to single modalities and thus fail to to exploit their complementary strengths. For instance, time-series models are suited for dynamics but miss spatial information, while image-based models capture global patterns but lose temporal details. This highlights the need for multimodal modeling, which faces several key challenges.

1) Heterogeneous Continuous–Discrete Multimodal Unified Representation. Time-series, images, and texts differ fundamentally in structure, semantics, and temporal continuity. Time-series data is continuous and high-frequency, whereas text is discrete and semantic. Forcing continuous time series into a text-like tokenized representation inevitably results in temporal dynamic information loss. Consequently, achieving a unified representation of continuous temporal-frequency dynamics and discrete textual semantics remains an open problem.

2) Multimodal Alignment Convergence Confliction of Foundation Model. During backpropagation, dominant modalities (typically time-series) converge faster and generate larger gradients, dictating the optimization direction. This imbalance suppresses the learning of other modalities (text and vision), resulting in overfitting of the dominant modalities and underfitting of auxiliary one, ultimately degrading the overall fusion performance, even worse than single modality.

To bridge this gap, we propose VLT, the first multimodal foundation model to jointly model temporal, visual, and textual information, without forcing continuous signals into discrete language tokens. The key insight is to introduce frequency-spectrum visual representations, acting as a bridge between industrial temporal signals and semantic reasoning, since frequency spectra naturally reveal critical fault characteristics such as harmonics, resonance bands, and degradation patterns. In this paradigm, each modality plays a distinct complementary role. Time-series encoding represents continuous temporal dynamics, frequency-domain visual representations capture spectral patterns, and text provide domain knowledge. 

VLT consists of three key components. VLT first employs a Time-aware Mixture of Experts (Time-MoE) to adaptively encode diverse temporal patterns. This is followed by a Frequency–Text Augmented Learner, which converts time-series signals into frequency-domain images and incorporates domain knowledge via textual embeddings, enabling joint semantic-frequency representation learning within the VLM. Finally, a time-centric multimodal fusion predictor dynamically integrates information from all modalities, ensuring robust forecasting even with imbalanced modalities. The main contributions of this work are as follows.

1) \textbf{\textit{New Perspective}}: We propose VLT, a multimodal foundation model paradigm that jointly models temporal signals, frequency-domain visual representations, and domain textual knowledge within a single architecture. This paradigm is designed to explicitly leverage the complementary strengths of heterogeneous modalities, enabling holistic health perception and prognostics. To the best of our knowledge, this is the first attempt to systematically integrate industrial signals, frequency image, and language modalities into a unified PHM foundation model, opening a new direction for multimodal industrial intelligence.

2) \textbf{\textit{New Architecture}}: We develop a  unified multimodal encoding representation architecture that addresses the fundamental heterogeneity between continuous temporal dynamics and discrete semantic representations. Specifically, a Time-aware Mixture-of-Experts (Time-MoE) is introduced to adaptively capture diverse temporal patterns, while a Frequency–Knowledge Augmented Learner transforms time-series signals into frequency-domain visual representations and enriches them with domain knowledge via textual embeddings. It effectively mitigates information loss across multi-modalities.  

3) \textbf{\textit{New Approach}}: We develop a time-centric multimodal gradient alignment mechanism to resolves cross-modal optimization conflicts. It introduces gradient convergence normalization and reliability-aware dynamic reweighting that prevents the dominance of fast-converging modalities, enabling stable model training among imbanlanced modalities.

4) We validate VLT on 11 industrial  tasks covering turbofan engines, batteries and bearing diagnosis. VLT consistently outperforms representative baselines under full-sample regression, few-shot diagnosis, and cross-domain settings.

\section{Related works}

\subsection{Deep Learning for PHM Tasks}

With multi-sensor time series data, deep learning has become a main solution for Prognostics and Health Management (PHM). It supports key tasks such as anomaly detection \cite{zhou2021novel}, fault diagnosis \cite{jin2023adaptive}, and remaining useful life (RUL) prediction \cite{ren2023dynamic}. Its main advantage is the ability to learn useful features directly from raw signals.

Recurrent Neural Networks (RNNs)\cite{jin2022bi,zhou2021novel,zheng2017lstm}, Convolutional Neural Networks (CNNs)\cite{wen2018cnnfd,chen2020tcnn,jiang2019multiscalecnn} and their variants, are widely used for sequential sensor data. They are effective for modeling temporal dependence and  capturing local patterns in degradation signals.
Hybrid models combine the strengths of different backbones. CNN-LSTM \cite{ouyang2024combined} and CNN-BiLSTM-AT \cite{yuan2025lithium} use CNNs for local feature extraction and recurrent networks for long-term dynamics. These models are useful for battery health estimation, where both short-term signal change and long-term aging trends matter. 
Data--model joint frameworks \cite{zhang2023data} further improved battery RUL prediction by combining learned features with degradation knowledge. More recently, Transformer-based models \cite{ren2023dynamic} use self-attention to model long-range dependence and cross-sensor interaction. Compared with recurrent models, they are more effective for complex temporal relations. In battery applications, \cite{ardakani2024narx} proposed a NARX Transformer for state-of-health prediction and showed better use of long-cycle aging information.

Overall, PHM models have evolved from recurrent and convolutional networks to attention-based architectures. However, most existing methods are task-specific and focus only on numerical sensor data. This limits their generalization under domain shift, few-label settings, and diverse industrial scenarios.

\subsection{Multimodal Large Models for Industrial Time Series}

With the rise of foundation models, researchers have started using Large Language Models (LLMs) for time-series analysis. Existing methods mainly follow two directions. The first treats time series as language-like sequences and uses pretrained LLMs directly\cite{wang2025collm,zhou2023one}. The second uses cross-modal alignment, converting time series into text or images that can be processed by pretrained models\cite{jin2024time, wang2025diagllm}.

Several studies explored how to transfer pretrained knowledge to time-series tasks. One-Fits-All \cite{zhou2023one} showed that frozen language models can serve as general backbones for forecasting, classification, and anomaly detection. PromptCast \cite{xue2023promptcast} formulated forecasting as prompt-based generation. Time-LLM \cite{jin2024time} aligned temporal patches with text prototypes through prompting. In parallel, native time-series foundation models such as Chronos \cite{ansari2024chronos}, Timer \cite{liu2024timer}, Sundial \cite{liu2025sundial}, and \cite{das2024decoder} showed strong zero-shot and transfer ability after large-scale pretraining.

Recent work has moved toward multimodal learning. ChatTime \cite{wang2025chattime} jointly models numerical and textual inputs in one framework. GPT4MTS \cite{jia2024gpt4mts} used textual side information to improve forecasting. In industrial settings, domain-guided systems \cite{chen2025domain} showed the value of expert knowledge for battery estimation. DiagLLM \cite{wang2025diagllm} further combined spectrum images and expert knowledge through efficient fine-tuning, but it does not retain a raw time-series branch for direct temporal modeling.

In PHM, researchers have also introduced large models for diagnosis and prognosis. Tao et al. \cite{tao2025llm} converted vibration signals into text descriptors for LLM-based fault diagnosis. The TimeGPT-based method in \cite{sun2025fine} used pretrained generation ability for battery health estimation. Predictive pretraining methods \cite{zhao2025predictive} further modeled long-term aging patterns with sparse attention. However, these methods still mainly rely on structured numerical inputs and often ignore changing working conditions.

These studies show the promise of large models for PHM. However, most current methods still rely on single modalities. They do not fully use the complementary information from time-series signals, images, and text, which limits robustness and adaptability in real industrial environments.

\section{VLT Architecture}

\begin{figure*}[!htb]
	\centering
	\includegraphics[width=2\columnwidth]{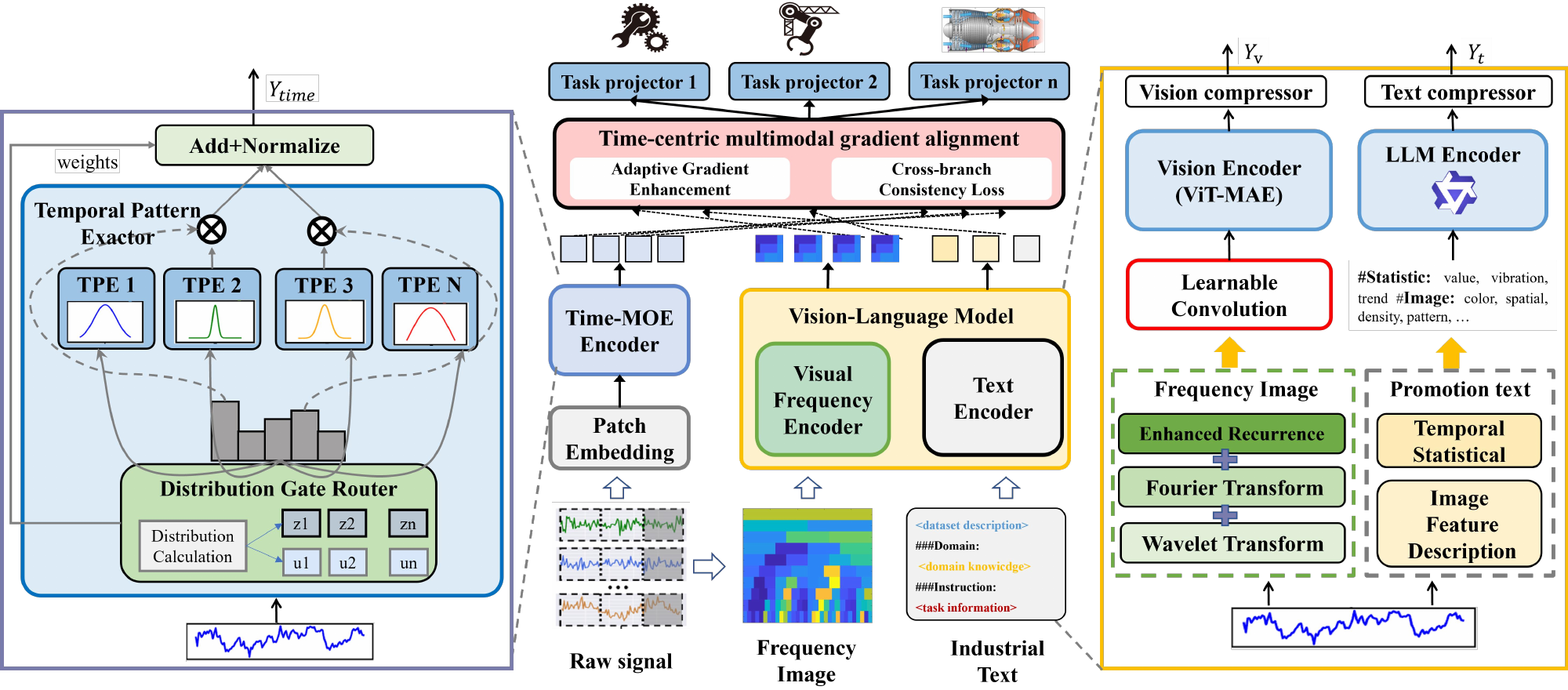}
	\caption{The overall architecture of the proposed VLT framework. The model integrates three modalities via dedicated branches: a Time-aware Mixture of Experts (Time-MoE) for raw time-series signals, a Visual Frequency Encoder for frequency-domain images, and a Text Encoder for domain-knowledge prompts. A time-centric multimodal gradient alignment mechanism is employed to dynamically balance modal contributions and align branch-level predictions prior to the final attention-based fusion.}
	\label{fig:architecture}
\end{figure*}

\subsection{Overall Architecture}
The overall architecture of VLT is a unified end-to-end framework, designed to integrate three different modalities, namely raw time-series data, frequency-domain visual representations, and textual domain knowledge. As shown in Figure \ref{fig:architecture}, the model is built on a pretrained Vision-Language Model (VLM) and contains three dedicated encoder branches.

For clarity, we use $X=[x_1;\,x_2;\,\ldots;\,x_B]\in \mathbb{R}^{B \times L \times C}$ to denote a batch of multivariate time-series windows, where $x_b\in\mathbb{R}^{L\times C}$ is the $b$-th sample, $B$ is the batch size, $L$ is the window length, and $C$ is the number of sensor channels. The three modalities are indexed by $m \in \mathcal{M}=\{\mathrm{T},\mathrm{V},\mathrm{K}\}$, where $\mathrm{T}$, $\mathrm{V}$, and $\mathrm{K}$ denote the temporal, visual, and knowledge branches, respectively. The corresponding embeddings are denoted as $\mathbf z_b^{\mathrm{T}}$, $\mathbf z_b^{\mathrm{V}}$, and $\mathbf z_b^{\mathrm{K}}$, and the final fused embedding is denoted as $\mathbf z_b^{\mathrm{F}}$.

The Time-aware Mixture of Experts (Time-MoE) branch processes the raw time-series signals and produces $\mathbf z_b^{\mathrm{T}}$. The Frequency Visual Learner converts the same time series into frequency-domain images and produces $\mathbf z_b^{\mathrm{V}}$. The Knowledge Learner encodes domain-specific text prompts and produces $\mathbf z_b^{\mathrm{K}}$. Finally, these modality embeddings are fused to generate the prediction $\hat y_b^{\mathrm{F}}$ for downstream PHM tasks.

\subsection{Time-aware Mixture of Experts (Time-MoE)}
Industrial time-series data are characterized by non-stationarity and heterogeneity. A single overall encoder often struggles to capture various dynamic behaviors, such as progressive degradation, sudden transient spikes, and quasi-periodic oscillations.

To achieve successful multimodal fusion, the temporal embedding $\mathbf z_b^{\mathrm{T}}$ must be robust and expressive. We therefore introduce a Time-MoE module, in which different experts specialize in different temporal patterns and a sparse router selects the most relevant experts for each sample.

For the $b$-th sample, the router computes expert logits and transforms them into expert probabilities as
\begin{equation}
\mathbf r_b = g(x_b) \in \mathbb{R}^{E}, \qquad
\mathbf p_b = \operatorname{softmax}(\mathbf W_h \mathbf r_b),
\end{equation}
where $E$ is the number of experts, $g(\cdot)$ is the routing network, $\mathbf W_h \in \mathbb{R}^{E \times E}$ is a learnable routing matrix, and $p_{b,e}$ denotes the $e$-th element of $\mathbf p_b$.

To reduce computation, only the top-$k_{\text{top}}$ experts are activated for each sample. Let $\mathcal S_b=\operatorname{TopK}(\mathbf p_b,k_{\text{top}})$ be the selected expert set. The sparse routing weight is defined as
\begin{equation}
a_{b,e}=
\begin{cases}
\dfrac{p_{b,e}}{\sum_{j\in \mathcal S_b} p_{b,j}}, & e\in \mathcal S_b,\\[6pt]
0, & e\notin \mathcal S_b.
\end{cases}
\end{equation}
Thus, $\sum_{e=1}^{E}a_{b,e}=1$ and unselected experts have zero routing weight.

Each expert $f_e(\cdot)$ is a lightweight temporal encoder. The Time-MoE output and the temporal embedding are computed:
\begin{equation}
\begin{aligned}
\mathbf h_b^{\mathrm{T}} &= \sum_{e=1}^{E} a_{b,e} f_e(x_b),\\
\mathbf z_b^{\mathrm{T}}
&=P_{\mathrm{T}}\left(\frac{1}{L}\sum_{t=1}^{L}
\mathbf h_{b,t}^{\mathrm{T}}\right),
\end{aligned}
\end{equation}
where $\mathbf h_{b,t}^{\mathrm{T}}$ denotes the hidden state at time step $t$, and $P_{\mathrm{T}}(\cdot)$ projects the averaged temporal representation into the common fusion dimension.

To avoid routing collapse where only a few experts are frequently selected, we use a balance regularization term. For expert $e$, the importance $u_e=\sum_{b=1}^{B}a_{b,e}$ measures its total routing weight, while the load $v_e$ is the number of samples selecting this expert.

\begin{equation}
\begin{aligned}
\mathcal{L}_{\text{bal}}
&= \operatorname{cv}^2(\mathbf u)+\operatorname{cv}^2(\mathbf v),\\
\operatorname{cv}^2(\mathbf q)
&= \frac{\operatorname{Var}(\mathbf q)}
{\operatorname{Mean}(\mathbf q)^2+\epsilon_{\operatorname{cv}}},
\end{aligned}
\end{equation}
where $\mathbf u=[u_1,\ldots,u_E]$ and $\mathbf v=[v_1,\ldots,v_E]$. This term encourages experts to have balanced routing weights and balanced sample loads.

\subsection{Frequency Visual Learner}

The time-series branch focuses on temporal dynamics but often struggles to capture subtle patterns in the frequency domain, which are crucial for diagnosing mechanical faults (e.g., bearing defects, gear meshing frequency anomalies).

The Frequency Visual Learner maps each time-series sample $x_b$ to a pseudo-color image $\mathbf I_b$, and then extracts a visual embedding $\mathbf z_b^{\mathrm{V}}$ using a pretrained visual encoder.

\noindent Time-to-Image Transformation Module.

First, recurrence information is computed to describe temporal self-similarity:
\begin{equation}
R_{b,i,j}
= \Theta\left(\epsilon_R-\left\|x_{b,i}-x_{b,j}\right\|_2\right),
\end{equation}
where $\Theta(\cdot)$ is the Heaviside step function, $\epsilon_R$ is the recurrence threshold, and $x_{b,i}$ denotes the sensor vector at time step $i$.

For compactness, let $x_{b,c}[n]$ denote the value of channel $c$ at time step $n$ in sample $b$.
Second, Fourier amplitudes are computed to capture global frequency components:
\begin{equation}
A_{b,c,k}
= \left|\sum_{n=0}^{L-1} x_{b,c}[n]\,
\exp\left(-\mathrm{j}2\pi kn/L\right)\right|,
\end{equation}
where $A_{b,c,k}$ is the amplitude of the $k$-th frequency component in channel $c$.

Third, wavelet coefficients are computed to describe local time-frequency patterns:
\begin{equation}
W_{b,c}(a,\tau)
= \sum_{n=0}^{L-1} x_{b,c}[n]\,\psi_{a,\tau}^{*}(n),
\end{equation}
where $a$ is the scale parameter, $\tau$ is the translation index, and $\psi_{a,\tau}^{*}(n)=a^{-1/2}\psi^*((n-\tau)/a)$ is the scaled and translated conjugate wavelet.

Let the feature maps generated from recurrence, Fourier, and wavelet branches be $\mathbf F^{\mathrm{R}}$, $\mathbf F^{\mathrm{A}}$, and $\mathbf F^{\mathrm{W}}$, respectively. A learnable image generator maps them to a pseudo-color image batch:
\begin{equation}
\mathbf I
= \Phi_{\mathrm{img}}\left(\mathbf F^{\mathrm{R}},\mathbf F^{\mathrm{A}},\mathbf F^{\mathrm{W}}\right)
\in\mathbb{R}^{B\times 3\times H_0\times W_0},
\end{equation}
where $\Phi_{\mathrm{img}}(\cdot)$ is implemented by lightweight convolutional layers. In our experiments, $H_0=W_0=112$.

We use a pretrained Masked Autoencoder (MAE) as the visual encoder. The visual embedding $\mathbf z_b^{\mathrm{V}}$ is obtained by feeding $\mathbf I_b$ into the MAE encoder $E_{\mathrm{V}}(\cdot)$ and projecting its output through $P_{\mathrm{V}}(\cdot)$. The last several MAE layers are selectively fine-tuned to adapt general visual representations to PHM tasks.

\subsection{Knowledge Learner}
Time-series and frequency representations describe degradation patterns at the signal level, but they provide limited semantic context. The Knowledge Learner introduces domain knowledge through natural-language prompts. These prompts may describe operating conditions, equipment information, or fault-related cues, such as load frequency, engine component type, and bearing fault location.

\noindent Cross-Modal Fusion Prompt Generation.
For both lifetime prediction and fault classification, the prompt combines statistical information from the raw time series with visual descriptions derived from the pseudo-color image.

Time-series Statistical Features. We extract statistics from three complementary aspects: baseline state, fault-sensitive fluctuation, and lifetime-related trend. The baseline state is described by the minimum, maximum, mean, and median values; fluctuation is described by standard deviation, kurtosis, and skewness; and trend is described by the global slope and average rate of change.

Visual Image Features. The pseudo-color image is summarized by color distribution, texture, spatial layout, and density patterns. These descriptions translate numerical degradation patterns into visual semantics that can be processed by the language encoder.

\noindent Lightweight Language Model Encoder.
For sample $b$, the text prompt $\mathbf t_b$ is constructed from time-series statistical features and visual descriptions. It is encoded by the Qwen1.5-0.5B text encoder $E_{\mathrm{K}}(\cdot)$ and projected by $P_{\mathrm{K}}(\cdot)$ to obtain the knowledge embedding $\mathbf z_b^{\mathrm{K}}$. Similar to the visual encoder, most language-model parameters are frozen and only the last several Transformer layers are fine-tuned.

The performance of multimodal fusion systems is often affected by modality imbalance. To address this issue, we use adaptive gradient adjustment and consistency constraints to balance modality contributions and align branch-level predictions.

\begin{figure*}[!htb]
    \centering
    \hspace*{-0mm}\includegraphics[width=0.8\linewidth]{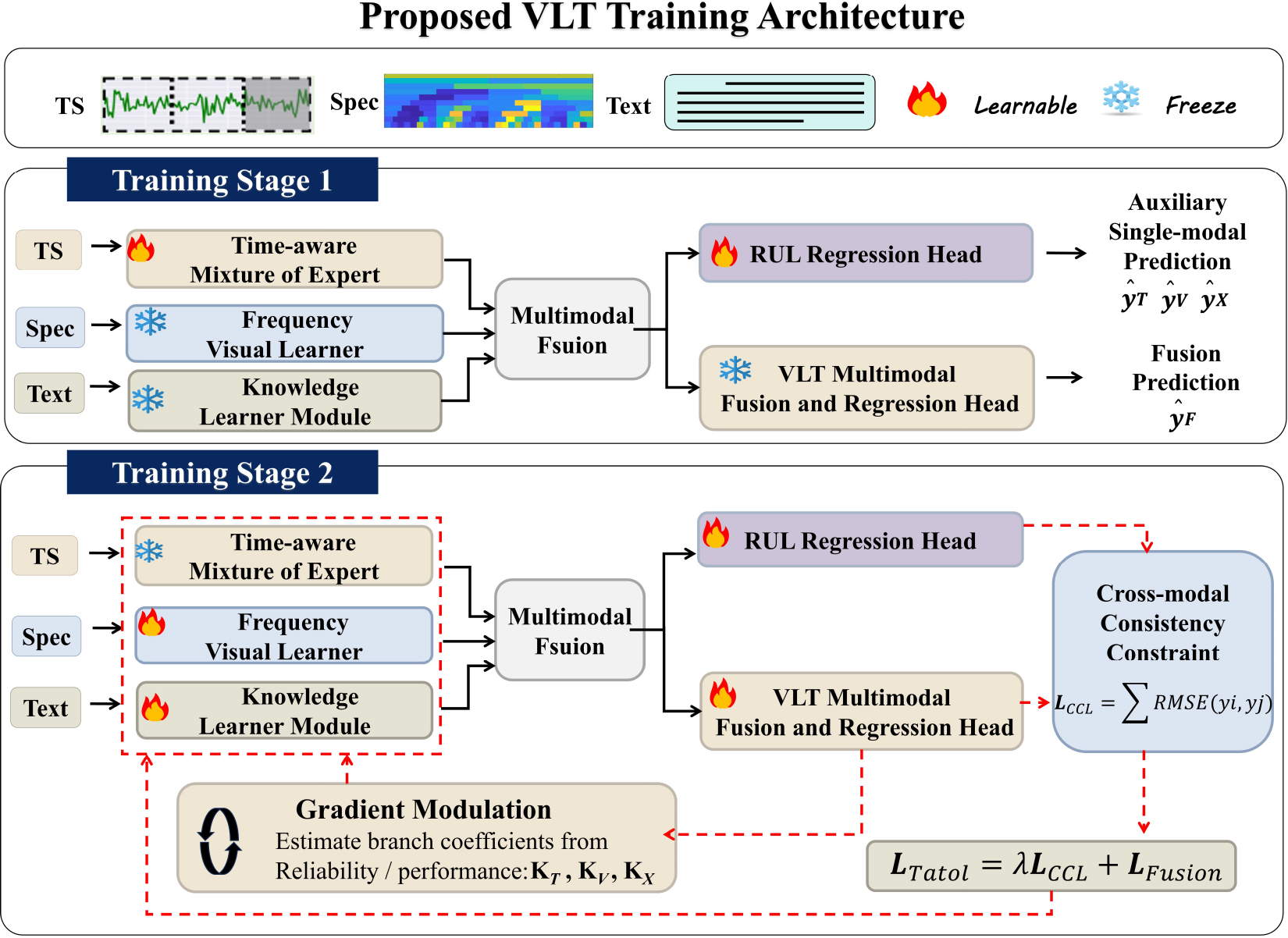}
    \caption{The proposed two-stage training architecture for multimodal imbalance learning. Stage 1 primarily optimizes the temporal experts, while Stage 2 fine-tunes the visual and knowledge branches guided by cross-branch consistency constraints and adaptive gradient modulation.}
    \label{fig:training_framework_v6}
\end{figure*}

\subsection{Time-centric multimodal gradient alignment mechanism}
The performance of multimodal fusion systems is easily constrained by the imbalance problem. We propose a multimodal imbalanced learning mechanism integrating adaptive gradient adjustment and consistency constraints. By dynamically optimizing modal contribution weights and aligning the prediction logic of each branch, balanced utilization of multimodal information is achieved. Fig.\ref{fig:training_framework_v6} illustrates the two-stage training architecture of VLT, where Stage 1 learns a stable temporal representation and Stage 2 fine-tunes the visual and knowledge branches with consistency constraints and adaptive gradient modulation.

The three modality embeddings are fused through modality attention and time-centered cross-modal attention. Let $e_b^m$ denote the learned attention score of modality $m$. The normalized modality weight and attention-weighted representation are computed as
\begin{equation}
\alpha_b^m=\frac{\exp(e_b^m)}
{\sum_{n\in\mathcal{M}}\exp(e_b^n)}, \qquad
\mathbf z_b^{\alpha}=\sum_{m\in\mathcal{M}}\alpha_b^m\mathbf z_b^m.
\end{equation}
where $\mathbf z_b^{\alpha}$ summarizes the contributions of all modality embeddings according to their learned weights. In parallel, a time-centered cross-modal representation is computed as
\begin{equation}
\mathbf z_b^{\mathrm{TC}}
=\mathcal A_{\mathrm{TC}}\left(\mathbf z_b^{\mathrm{T}},
\{\mathbf z_b^m\}_{m\in\mathcal{M}}\right),
\end{equation}
where $\mathcal A_{\mathrm{TC}}(\cdot)$ uses the temporal embedding as the anchor to aggregate information from all modalities. The final fused embedding is
\begin{equation}
\mathbf z_b^{\mathrm{F}}=P_{\mathrm{F}}\left(
\left[\mathbf z_b^{\alpha};\mathbf z_b^{\mathrm{TC}}\right]\right).
\end{equation}
Here, $P_{\mathrm{F}}(\cdot)$ is the fusion projection layer, $[\,;\,]$ denotes vector concatenation, and $\mathrm{TC}$ denotes time-centered. Each modality branch and the fused representation then produce their predictions through lightweight prediction heads.

To address the prediction confidence differences between various modality branches, we design the Adaptive Gradient Enhancement (AGE) mechanism, which dynamically adjusts the gradient update magnitude of each branch, allowing weaker modalities to receive more thorough training.

\begin{algorithm}[t]
\small
\SetAlgoNlRelativeSize{0}
\DontPrintSemicolon
\caption{Time-Centric Training Procedure}
\label{alg:time_centric}

\KwIn{A mini-batch of multimodal samples}
\KwOut{Updated parameters of VLT}

Extract embeddings from the temporal, visual, and knowledge branches;\;
Fuse the branch embeddings with modality attention;\;

Use the temporal embedding as the anchor:
\[
\mathbf z_b^{\mathrm{TC}}
=\mathcal A_{\mathrm{TC}}\left(
\mathbf z_b^{\mathrm{T}},
\{\mathbf z_b^m\}_{m\in\mathcal{M}}
\right)\,;
\]\;

Project the attended and time-centered features into the fused embedding;\;
Generate branch-level predictions and the fused prediction;\;
Compute branch losses, fusion loss, modality confidence scores,
adaptive coefficients, and the CCL term;\;

Form the training objective:
\[
\mathcal L_{\text{total}}
=\mathcal L_{\mathrm{F}}
+\eta\sum_{m\in\mathcal{M}}\mathcal L_m
+\lambda_{\mathrm{bal}}\mathcal L_{\mathrm{bal}}
+\lambda_{\mathrm{c}}\mathcal L_{\mathrm{CCL}}\,;
\]\;

Back-propagate $\mathcal L_{\text{total}}$ to obtain branch gradients
$\mathbf g_m$;\;

Apply time-centric gradient normalization:
\[
\tilde{\mathbf g}_m=
\begin{cases}
\mathbf g_{\mathrm{T}}, & m=\mathrm{T},\\[4pt]
c_m
\dfrac{\|\mathbf g_{\mathrm{T}}\|_2}
{\|\mathbf g_m\|_2+\epsilon_g}
\mathbf g_m,
& m\in\{\mathrm{V},\mathrm{K}\}
\end{cases}
\,;
\]\;

Update branch parameters using $\tilde{\mathbf g}_m$;\;

\end{algorithm}

To quantify the reliability of each modality branch, the Confidence Score Calculation step first evaluates the prediction losses of both individual branches and the fused output. Let $\ell(\cdot,\cdot)$ be the task loss, which is MSE for regression and cross-entropy for classification. Let $\hat y_b^m$, $\hat y_b^{\mathrm{F}}$, and $y_b$ denote the prediction of branch $m$, the fused prediction, and the ground-truth target, respectively. The branch and fusion losses are
\begin{equation}
\begin{aligned}
\mathcal L_m&=\frac{1}{B}\sum_{b=1}^{B}
\ell(\hat y_b^m,y_b), \qquad
\mathcal L_{\mathrm{F}}&=\frac{1}{B}\sum_{b=1}^{B}
\ell(\hat y_b^{\mathrm{F}},y_b).
\end{aligned}
\end{equation}
The confidence scores are defined as
\begin{equation}
s_m=\frac{1}{1+\mathcal L_m}, \qquad
s_{\mathrm{F}}=\frac{1}{1+\mathcal L_{\mathrm{F}}}.
\end{equation}
Higher $s_m$ indicates a more reliable modality branch.

Based on the obtained confidence scores, the Gradient Adaptive Adjustment step assigns a larger gradient coefficient to weaker branches so that less reliable modalities can receive stronger updates. Therefore, the reliability-aware coefficient is defined as
\begin{equation}
c_m=1+\gamma
\frac{[s_{\mathrm{F}}-s_m]_+}{s_{\mathrm{F}}+\epsilon_g}.
\end{equation}
Here, $[x]_+=\max(x,0)$, $\gamma$ controls the enhancement strength, and $\epsilon_g$ is a numerical stability constant. Let $\theta_m$ denote the parameters of branch $m$, and let $\mathbf g_m=\nabla_{\theta_m}\mathcal L_{\text{total}}$ be its gradient. The time-centric normalized gradient is
\begin{equation}
\tilde{\mathbf g}_m=
\begin{cases}
\mathbf g_{\mathrm{T}}, & m=\mathrm{T},\\[4pt]
c_m\dfrac{\|\mathbf g_{\mathrm{T}}\|_2}{\|\mathbf g_m\|_2+\epsilon_g}\mathbf g_m,
& m\in\{\mathrm{V},\mathrm{K}\}.
\end{cases}
\end{equation}
The optimizer updates each branch using $\tilde{\mathbf g}_m$. This normalizes auxiliary-branch gradients with respect to the temporal branch and gives additional updates to low-confidence modalities.

To further mitigate modality bias caused by skewed sample distributions, we introduce the Cross-Branch Consistency Constraint (CCL) to enforce consistent prediction logic across modality branches. The constraint is applied to the temporal--visual, temporal--knowledge, and visual--knowledge modality pairs. For $(m,n)\in\mathcal P$, where $\mathcal P$ denotes this set of modality pairs, the reliability weight is
\begin{equation}
w_{mn}=\frac{s_ms_n}{\sum_{(p,q)\in\mathcal P}s_ps_q},
\quad (m,n)\in\mathcal P.
\end{equation}
The total consistency loss is
\begin{equation}
  \mathcal{L}_{\text{CCL}}
  = \sum_{(m,n)\in\mathcal P}w_{mn}
  \frac{1}{B}\sum_{b=1}^{B}
  D(\hat y_b^m,\hat y_b^n),
\end{equation}
where $D(\cdot,\cdot)$ is MSE for regression and KL or JS divergence for classification. The final training objective is
\begin{equation}
  \mathcal L_{\text{total}}
  =\mathcal L_{\mathrm{F}}+\eta\sum_{m\in\mathcal{M}}\mathcal L_m
  +\lambda_{\mathrm{bal}}\mathcal L_{\text{bal}}
  +\lambda_{\mathrm{c}}\mathcal L_{\text{CCL}},
\end{equation}
where $\eta$, $\lambda_{\mathrm{bal}}$, and $\lambda_{\mathrm{c}}$ control branch supervision, expert balance, and cross-branch consistency, respectively.

The training procedure of the proposed time-centric mechanism is summarized
in Algorithm~\ref{alg:time_centric}.

In summary, the proposed multimodal imbalance learning mechanism combines AGE, CCL, and attention fusion to enhance weak modalities, align branch-level predictions, and optimize modality weights. This design provides a robust feature fusion foundation for equipment lifetime prediction and fault classification tasks.

\section{Experimental Results}
\subsection{Experimental Setup}

\subsubsection{Datasets}
We conduct experiments on three public industrial benchmark datasets, covering engine RUL prediction, battery health estimation, and bearing fault diagnosis. Detailed dataset statistics and analysis are provided in the supplementary material (Tables S1-S3); here we briefly summarize their roles in the experiments.The main hyperparameter settings are summarized in Table~\ref{tab:pred_hparams}, including optimization parameters, input window size, Time-MoE configurations, and spectrum-aware vision-language module settings.

\begin{table}[t]
\centering
\caption{\textsc{Hyperparameter Settings}}
\label{tab:pred_hparams}
\begin{tabular}{ccc} 
\toprule
Category & Parameter & Value \\
\midrule

% --- Optimization ---
\multirow{3}{*}{Optimization} 
 & Batch Size & 128 \\
 & Learning Rate & $1\times10^{-3}$ \\
 & Total Epochs & 130 \\
\midrule

% --- Data Input ---
\multirow{2}{*}{Data Input} 
 & Window Size ($L$) & 48 \\
 & Prediction Length & 1 \\
\midrule

% --- Temporal Branch ---
\multirow{8}{*}{Temporal Branch} 
 & Patch Length ($P$) & 16 \\
 & Patch Stride ($S$) & 8 \\
 & Model Dim ($d_{model}$) & 48 \\
 & Hidden Dim & 64 \\
 & Attention Heads & 1 \\
 & Number of Experts & 6 \\
 & Top-$k$ Experts & 3 \\
 & Dropout Rate & 0.1 \\
\midrule

% --- Spectrum-aware VLMA ---
\multirow{4}{*}{Spectrum-aware VLMA} 
 & Image Resolution & $112\times112$ \\
 & Visual Embed Dim & 128 \\
 & Text Feature Dim & 96 \\
 & Max Token Length & 512 \\
\bottomrule
\end{tabular}
\end{table}

\textbf{C-MAPSS Dataset:} We use the C-MAPSS turbofan benchmark for engine RUL prediction, covering four subsets with different operating conditions and fault modes.

\textbf{XJTU Battery Dataset:} We use six representative XJTU battery subsets to evaluate capacity degradation prediction under different charge-discharge protocols.

\textbf{CWRU Bearing Dataset:} We construct a 10-class bearing fault diagnosis task from the CWRU drive-end vibration signals to evaluate few-shot classification.

\subsubsection{Evaluation Metrics}
To ensure a fair and task-appropriate assessment, we adopt the following evaluation metrics for different task types.

\textbf{Regression Tasks.} Root Mean Square Error (RMSE) and the asymmetric Score function serve as the primary metrics. RMSE measures the overall prediction accuracy, defined as
\begin{equation}
\text{RMSE} = \sqrt{\frac{1}{N}\sum_{i=1}^{N}(\hat{y}_i - y_i)^2},
\end{equation}
where $N$ is the number of samples, $\hat{y}_i$ and $y_i$ denote the predicted and true RUL of the $i$-th sample, respectively. A lower RMSE indicates better performance.

The score uses an unequal standard. Late predictions receive stricter punishment, as overrating RUL brings greater risks in practice.
\begin{equation}
\text{Score} =
\begin{cases}
\sum_{i=1}^{N}\left[\exp\left(-\dfrac{\hat{y}_i - y_i}{13}\right) - 1\right], & \hat{y}_i < y_i, \\[6pt]
\sum_{i=1}^{N}\left[\exp\left(\dfrac{\hat{y}_i - y_i}{10}\right) - 1\right], & \hat{y}_i \geq y_i.
\end{cases}
\end{equation}
Lower Score values indicate better predictive performance.

\textbf{Classification Tasks.}  Accuracy measures the overall classification correctness.
\begin{equation}
\mathrm{Accuracy} =
\frac{TP+TN}{TP+TN+FP+FN},
\end{equation}
where the four terms denote the counts of true-positive, true-negative, false-positive, and false-negative predictions. To complement accuracy, the F1 score is further used, which combines precision and recall into a single harmonic-mean metric and therefore offers a more robust criterion for imbalanced classification tasks.

\begin{equation}
\text{F1} = \frac{2TP}{2TP + FP + FN}.
\end{equation}

\subsection{Regression Tasks Results}
To evaluate the performance of VLT on industrial time series prediction tasks, we conduct full-sample and few-shot experiments on the C-MAPSS and XJTU battery datasets respectively in this section, and compare it with representative time series forecasting methods including Time-LLM~\cite{jin2024time}, GPT4TS~\cite{zhou2023one}, DLinear~\cite{zeng2023transformers}, PatchTST~\cite{nie2023patchtst}, and TimesNet~\cite{wu2023timesnet}.

\subsubsection{Full-sample Training Results}

\begin{table*}[!t]
\centering
\setlength{\tabcolsep}{4pt}
\caption{Regression Task Results on the Full Training Dataset}
\label{tab:full_sample_results_100}
\begin{tabular}{llcccccccccccc}
\toprule
\multirow{2}{*}{Category} & \multirow{2}{*}{Dataset}
& \multicolumn{2}{c}{VLT}
& \multicolumn{2}{c}{Time-LLM~\cite{jin2024time}}
& \multicolumn{2}{c}{GPT4TS~\cite{zhou2023one}}
& \multicolumn{2}{c}{DLinear~\cite{zeng2023transformers}}
& \multicolumn{2}{c}{PatchTST~\cite{nie2023patchtst}}
& \multicolumn{2}{c}{TimesNet~\cite{wu2023timesnet}} \\ 
\cmidrule(lr){3-4} \cmidrule(lr){5-6} \cmidrule(lr){7-8} \cmidrule(lr){9-10} \cmidrule(lr){11-12} \cmidrule(lr){13-14}
& & RMSE & Score & RMSE & Score & RMSE & Score & RMSE & Score & RMSE & Score & RMSE & Score \\ 
\midrule

\multirow{4}{*}{\textbf{C-MAPSS}}
& FD001
& \textbf{11.47} & \textbf{303.70}
& 13.51 & 270.10
& 12.68 & 261.80
& 13.23 & 307.50
& \underline{12.56} & \underline{235.80}
& 14.65 & 359.30 \\

& FD002
& \textbf{15.50} & \textbf{1441.08}
& 42.85 & 36080.00
& 26.68 & 24947.66
& \underline{22.90} & \underline{5611.24}
& 26.66 & 9634.02
& 23.25 & 8002.84 \\

& FD003
& \textbf{11.15} & \textbf{214.40}
& 19.59 & 876.00
& 13.29 & 392.10
& 14.21 & 428.20
& \underline{12.59} & \underline{390.60}
& 13.68 & 515.90 \\

& FD004
& \textbf{16.59} & \textbf{1556.20}
& 42.94 & 44291.06
& 31.19 & 32235.04
& 27.54 & 7922.36
& 30.47 & 14822.96
& \underline{26.81} & \underline{12782.66} \\

\addlinespace[3pt]

\multirow{6}{*}{\textbf{Battery ($\times 10^{-1}$)}}
& 2C
& \textbf{0.12} & \textbf{2.57}
& 0.30 & 5.72
& \underline{0.20} & \underline{2.86}
& 0.42 & 8.58
& 0.27 & 5.72
& 0.21 & 2.86 \\

& 3C
& \underline{0.14} & \underline{2.93}
& 0.21 & 6.50
& \textbf{0.13} & \textbf{3.25}
& 0.18 & 3.25
& 0.20 & 3.25
& \textbf{0.13} & \textbf{3.25} \\

& R2.5
& \textbf{0.07} & \textbf{1.71}
& 0.29 & 8.56
& 0.13 & 4.28
& \underline{0.09} & \underline{4.28}
& 0.21 & 8.56
& 0.13 & 4.28 \\

& R3
& \underline{0.09} & \underline{3.54}
& 0.26 & 8.86
& 0.16 & 4.43
& \textbf{0.08} & \textbf{4.43}
& 0.18 & 4.43
& 0.20 & 4.43 \\

& RW
& \textbf{0.16} & \textbf{0.83}
& 0.18 & 1.39
& \underline{0.15} & \underline{1.39}
& 0.31 & 2.78
& 0.24 & 2.78
& 0.18 & 1.39 \\

& Satellite
& \textbf{0.06} & \textbf{2.24}
& 0.21 & 5.61
& 0.12 & 5.61
& \underline{0.09} & \underline{5.61}
& 0.13 & 5.61
& \underline{0.09} & \underline{5.61} \\

\bottomrule
\end{tabular}

\end{table*}
\begin{figure*}
    \centering
    \includegraphics[width=1\linewidth]{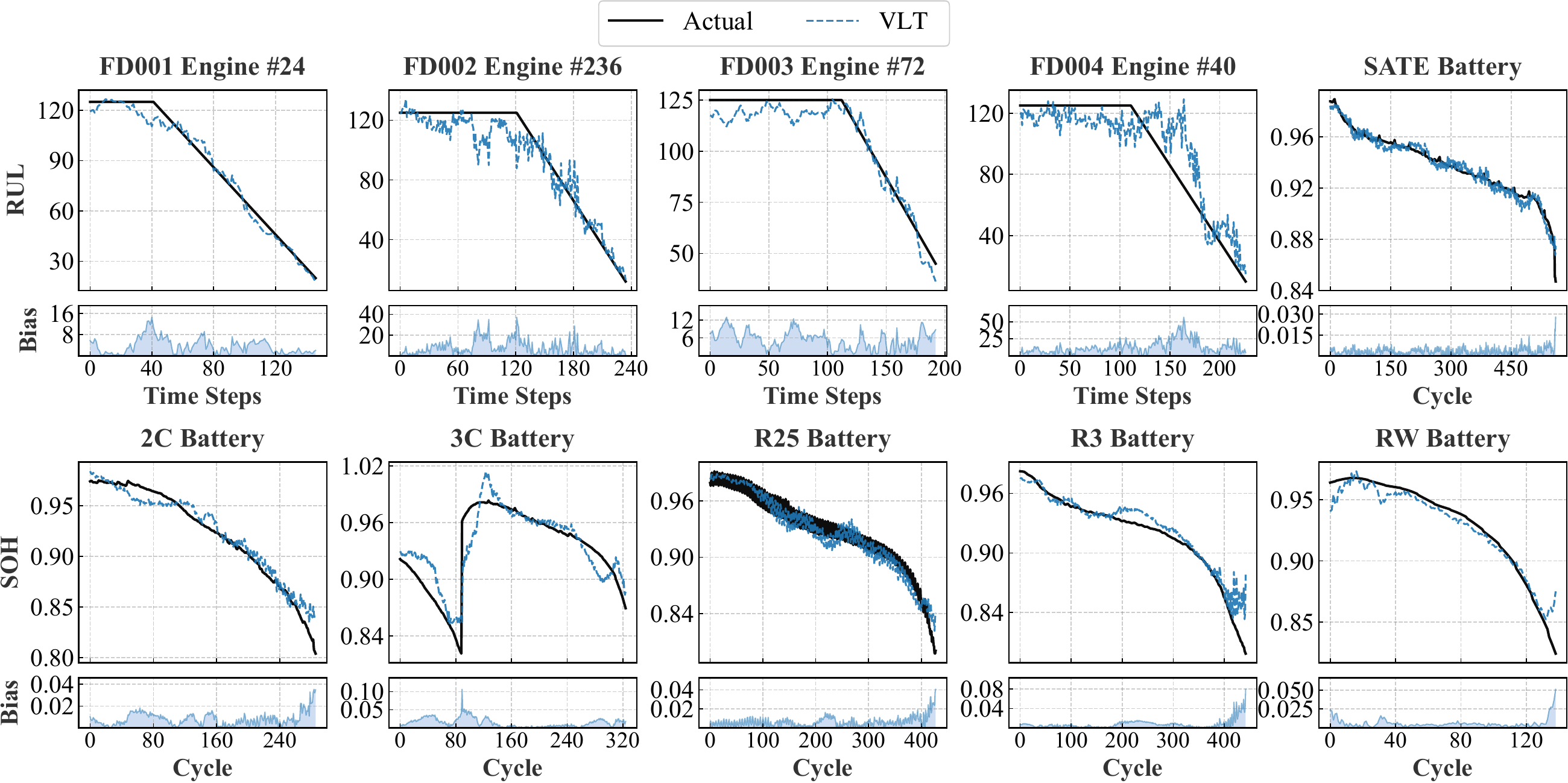}
    \caption{Qualitative regression visualization on the C-MAPSS engine and XJTU battery datasets. Representative testing units from FD001--FD004 are reported for C-MAPSS, and typical capacity-fading trajectories are shown for the battery dataset. For each subset, the upper panel compares predictions with ground truth, while the lower panel depicts the absolute prediction error over time.}
    \label{fig:Visualization of several prediction results}
\end{figure*}

As shown in Table~\ref{tab:full_sample_results_100}, VLT achieves the lowest RMSE values on all four C-MAPSS subsets, demonstrating the best overall prediction performance under the RMSE-prioritized comparison protocol. On FD001, its RMSE is 11.47, approximately 8.7\% lower than that of the second-best method PatchTST. On the more challenging multi-operating-condition subsets FD002 and FD004, the advantages of VLT are even more prominent: its RMSE is much lower than those of DLinear, GPT4TS, and Time-LLM, and its Score is substantially lower than most baselines. It also achieves optimal results on FD003 and FD004.

In the battery capacity estimation task, VLT delivers the best or near-best performance on most subsets. On the 2C and R2.5 subsets, its RMSE is significantly superior to those of Time-LLM and DLinear, accompanied by a sharp reduction in Score, and it maintains optimal performance on the RW and Satellite subsets. Only on the 3C and R3 subsets is its RMSE slightly inferior to individual comparison models, but overall it achieves favorable comprehensive performance under various battery operating conditions.

Comparative experiments show that the large language model-based methods Time-LLM and GPT4TS perform adequately only on the simple FD001 subset, while their errors increase drastically in complex multi-operating-condition scenarios, indicating that relying solely on language model-based temporal encoding is insufficient to characterize degradation features. In contrast, VLT exhibits stronger robustness in both aero-engine and multi-condition battery degradation prediction tasks.

As can be observed from the qualitative results in Figure~\ref{fig:Visualization of several prediction results}, the prediction curves of VLT are highly consistent with the real degradation trends, and the errors remain small near the failure point, showing stable trend-tracking and prediction performance.

In summary, both quantitative and qualitative experiments verify that VLT achieves higher accuracy and stronger stability in engine remaining useful life prediction and battery capacity estimation. By integrating frequency-domain visual information and textual domain knowledge, the proposed multimodal framework can model the degradation process more effectively and presents favorable generalization ability in multi-operating-condition degradation scenarios.

\subsubsection{Few-shot Learning Results}

\begin{table*}[!htb]
\centering
\setlength{\tabcolsep}{4pt}
\caption{Regression task results under few-shot learning (5\% training data)}
\label{tab:fewshot_5_results}
\begin{tabular}{llcccccccccccc}
\toprule
\multirow{2}{*}{Category} & \multirow{2}{*}{Dataset}
& \multicolumn{2}{c}{VLT}
& \multicolumn{2}{c}{Time-LLM~\cite{jin2024time}}
& \multicolumn{2}{c}{GPT4TS~\cite{zhou2023one}}
& \multicolumn{2}{c}{DLinear~\cite{zeng2023transformers}}
& \multicolumn{2}{c}{PatchTST~\cite{nie2023patchtst}}
& \multicolumn{2}{c}{TimesNet~\cite{wu2023timesnet}} \\ 
\cmidrule(lr){3-4} \cmidrule(lr){5-6} \cmidrule(lr){7-8} \cmidrule(lr){9-10} \cmidrule(lr){11-12} \cmidrule(lr){13-14}
& & RMSE & Score & RMSE & Score & RMSE & Score & RMSE & Score & RMSE & Score & RMSE & Score \\ 
\midrule

\multirow{4}{*}{\textbf{C-MAPSS}}
& FD001
& \textbf{13.75} & \textbf{355.90}
& 22.75 & 1393.50
& 16.09 & 456.20
& 30.83 & 3437.20
& \underline{14.91} & \underline{318.30}
& 15.17 & 288.20 \\

& FD002
& \textbf{26.07} & \textbf{6212.12}
& 43.07 & 45980.27
& 43.04 & 33563.29
& 43.87 & 46291.59
& 43.24 & 29030.79
& \underline{42.79} & \underline{29452.70} \\

& FD003
& \underline{12.97} & \underline{365.50}
& 17.59 & 550.50
& 15.84 & 813.70
& 39.62 & 22617.50
& \textbf{12.83} & \textbf{317.40}
& 15.06 & 573.70 \\

& FD004
& \textbf{27.57} & \textbf{12101.16}
& 42.99 & 40749.62
& \underline{42.67} & \underline{44701.50}
& 43.86 & 69222.26
& 43.30 & 60791.74
& 42.95 & 44170.29 \\

\addlinespace[3pt]

\multirow{6}{*}{\textbf{Battery ($\times 10^{-1}$)}}
& 2C
& \underline{0.41} & \underline{8.58}
& 0.46 & 8.58
& \textbf{0.36} & \textbf{5.72}
& 0.99 & 22.88
& 0.46 & 8.58
& 0.80 & 14.30 \\

& 3C
& \textbf{0.21} & \textbf{9.75}
& 0.40 & 9.75
& \underline{0.24} & \underline{6.50}
& 0.72 & 16.25
& 0.44 & 9.75
& 0.64 & 13.00 \\

& R2.5
& \textbf{0.23} & \textbf{8.56}
& 0.39 & 12.84
& 0.37 & 12.84
& \underline{0.26} & \underline{25.68}
& 0.46 & 12.84
& 0.56 & 12.84 \\

& R3
& \textbf{0.38} & \textbf{8.86}
& \underline{0.39} & \underline{22.15}
& 0.43 & 13.29
& 0.42 & 13.29
& 0.43 & 13.29
& 0.46 & 13.29 \\

& RW
& \textbf{0.40} & \textbf{4.17}
& \underline{0.47} & \underline{5.56}
& \textbf{0.40} & \textbf{4.17}
& 0.74 & 6.95
& 1.01 & 9.73
& 1.07 & 11.12 \\

& Satellite
& \textbf{0.23} & \textbf{11.22}
& 0.34 & 11.22
& \underline{0.25} & \underline{11.22}
& 0.28 & 11.22
& 0.27 & 11.22
& 0.44 & 16.83 \\

\bottomrule
\end{tabular}
\end{table*}

\begin{figure*}
    \centering
    \includegraphics[width=1\linewidth]{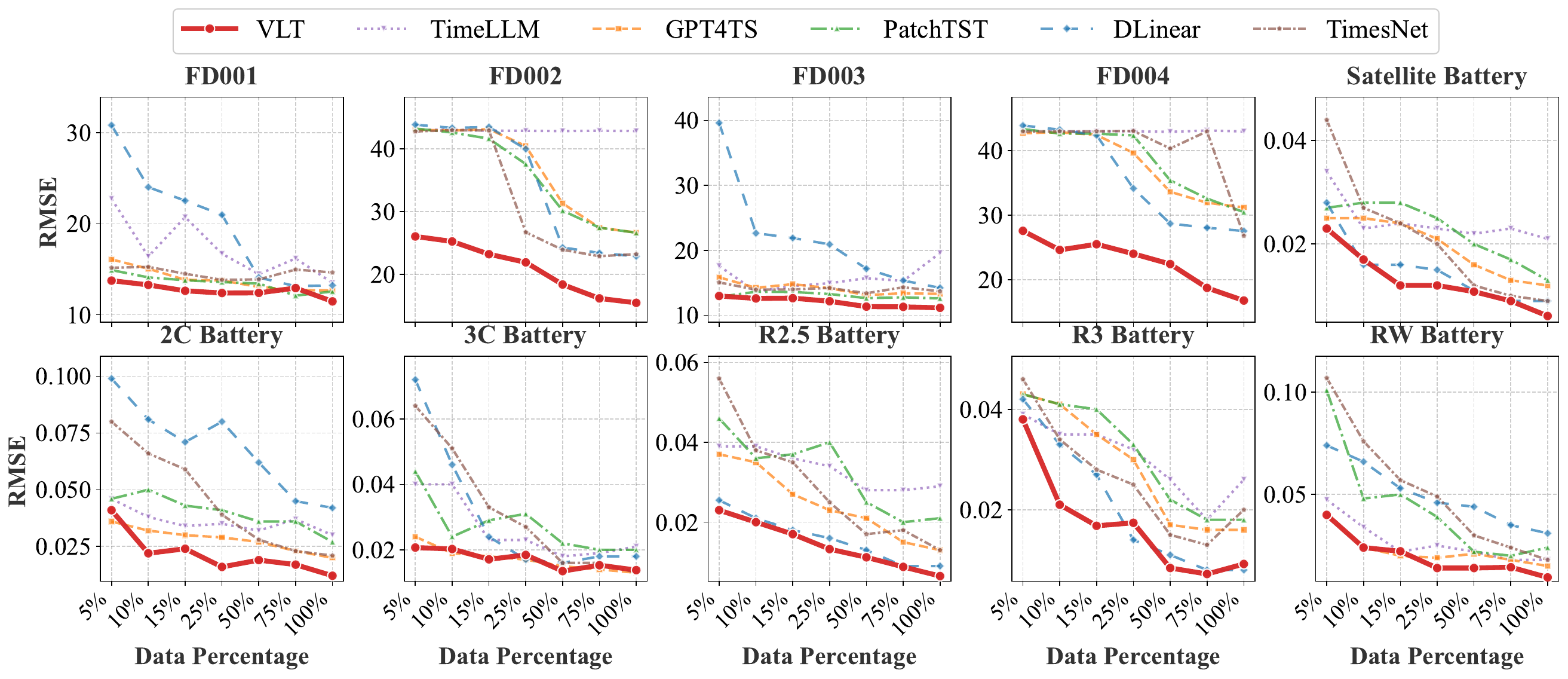}
    \caption{Performance of regression tasks under few-shot learning. The x-axis shows the percentage of training data used, from 5\% to 100\%. The y-axis denotes Root Mean Square Error, where lower is better.}
    \label{fig:Regression task results under few-shot learning}
\end{figure*}

To evaluate the few-shot generalization capability of VLT, we examine its performance under training data proportions ranging from 5 percent to 100 percent. In practical applications, labeled degradation data are often scarce, making this evaluation particularly important.

Table~\ref{tab:fewshot_5_results} shows that under 5 percent training data, VLT achieves an RMSE of 13.75 on FD001, outperforming the 22.75 of Time-LLM and the 30.83 of DLinear. The most substantial difference appears on the multi-operating-condition subsets. On FD002, VLT records an RMSE of 26.07 and is the only method with a value below 40, whereas the other baselines remain in the range from 42 to 44 and all yield Score values above 29000. A similar pattern is observed on FD004, where VLT attains 27.57, while all baseline methods remain above 42. 

%When the data budget increases to 10 percent, as reported in Table~\ref{tab:fewshot_10_results}, the gains achieved by VLT are substantially greater than those of the baselines. On FD001, the RMSE of VLT decreases from 13.75 to 13.28, approaching the 11.47 achieved under full-data training. On FD004, the RMSE of VLT declines from 27.57 to 24.62, and the Score decreases from 12101.16 to 5768.23, corresponding to a reduction of 52 percent. This result indicates more robust prediction performance, while the improvements of the other baseline methods remain limited. A consistent trend is also observed in the battery task, where the RMSE of VLT on 2C\_battery decreases from 0.41 to 0.22 as the amount of training data increases.

The data-efficiency curves from 5 percent to 100 percent shown in Figure~\ref{fig:Regression task results under few-shot learning} indicate that VLT shows clear advantages on the more complex FD002 and FD004 subsets and declines smoothly as more training data are used, whereas the baselines are nearly flat in the low-data region on these subsets. This suggests that multimodal fusion provides richer gradient signals and helps prevent the optimization process from converging to inferior solutions.

In summary, under data-scarce conditions, the prediction performance of conventional models deteriorates significantly on complex subsets. In contrast, VLT alleviates performance degradation by leveraging transferable vision-language representations and complementary multimodal information.

\begin{table*}[!t]
\centering
\caption{Fault classification task few-shot performance comparison measured by accuracy}
\label{tab:fewshot_test_acc}
\begin{tabular}{lccccc}
\toprule
\multirow{2}{*}{Methods} 
& \multicolumn{5}{c}{Accuracy} \\
\cmidrule(lr){2-6}
& 1-shot & 2-shot & 4-shot & 8-shot & Full Data \\
\midrule

VLT 
& \textbf{88.23 $\pm$ 0.59\%} 
& \textbf{93.91 $\pm$ 0.54\%} 
& \textbf{94.36 $\pm$ 0.29\%} 
& \textbf{94.53 $\pm$ 0.49\%} 
& 99.33 $\pm$ 0.18\% \\

FD-MVLLM~\cite{li2025fdmvllm}
& \underline{81.12 $\pm$ 1.21\%} 
& \underline{85.84 $\pm$ 0.65\%} 
& \underline{88.75 $\pm$ 0.67\%} 
& 89.94 $\pm$ 0.62\% 
& 95.26 $\pm$ 2.47\% \\

LiteFormer~\cite{sun2024liteformer}
& 36.40 $\pm$ 3.60\% 
& 54.92 $\pm$ 4.68\% 
& 68.41 $\pm$ 4.33\% 
& 86.06 $\pm$ 1.63\% 
& \underline{99.86 $\pm$ 0.25\%} \\

CNN\_LSTM~\cite{chen2021bearing}
& 19.24 $\pm$ 4.12\% 
& 22.60 $\pm$ 2.54\% 
& 35.83 $\pm$ 5.32\% 
& 59.92 $\pm$ 6.01\% 
& \textbf{100.00 $\pm$ 0.00\%} \\

DP\_MRTN~\cite{chen2022dualpath}
& 59.17 $\pm$ 13.11\% 
& 80.97 $\pm$ 2.65\% 
& 85.37 $\pm$ 1.82\% 
& 87.08 $\pm$ 1.29\% 
& \textbf{100.00 $\pm$ 0.00\%} \\

InversePINN~\cite{qin2024inversepinn}
& 51.91 $\pm$ 5.37\% 
& 70.19 $\pm$ 10.20\% 
& 82.80 $\pm$ 3.34\% 
& 88.52 $\pm$ 2.36\% 
& \textbf{100.00 $\pm$ 0.00\%} \\

TARTDN~\cite{cui2024tartdn}
& 66.94 $\pm$ 4.94\% 
& 66.95 $\pm$ 6.47\% 
& 73.33 $\pm$ 10.90\% 
& \underline{90.61 $\pm$ 3.22\%} 
& 98.99 $\pm$ 1.39\% \\

\bottomrule
\end{tabular}
\end{table*}

\begin{table*}[!t]
\centering
\caption{Fault classification task few-shot performance comparison measured by F1 score}
\label{tab:fewshot_test_f1}
\begin{tabular}{lccccc}
\toprule
\multirow{2}{*}{Methods} 
& \multicolumn{5}{c}{F1 Score} \\
\cmidrule(lr){2-6}
& 1-shot & 2-shot & 4-shot & 8-shot & Full Data \\
\midrule

VLT 
& \textbf{88.23 $\pm$ 0.58\%} 
& \textbf{93.91 $\pm$ 0.54\%} 
& \textbf{94.36 $\pm$ 0.29\%} 
& \textbf{94.53 $\pm$ 0.48\%} 
& 99.17 $\pm$ 0.39\% \\

FD-MVLLM~\cite{li2025fdmvllm}
& \underline{81.11 $\pm$ 1.21\%} 
& \underline{85.84 $\pm$ 0.65\%} 
& \underline{88.75 $\pm$ 0.67\%} 
& \underline{89.94 $\pm$ 0.62\%} 
& 95.39 $\pm$ 2.18\% \\

LiteFormer~\cite{sun2024liteformer}
& 31.66 $\pm$ 4.58\% 
& 53.17 $\pm$ 4.99\% 
& 65.78 $\pm$ 4.41\% 
& 85.36 $\pm$ 1.58\% 
& \underline{99.84 $\pm$ 0.28\%} \\

CNN\_LSTM~\cite{chen2021bearing}
& 15.62 $\pm$ 3.22\% 
& 19.96 $\pm$ 2.86\% 
& 33.78 $\pm$ 5.50\% 
& 59.26 $\pm$ 4.42\% 
& \textbf{100.00 $\pm$ 0.00\%} \\

DP\_MRTN~\cite{chen2022dualpath}
& 52.09 $\pm$ 15.38\% 
& 78.66 $\pm$ 2.22\% 
& 85.28 $\pm$ 1.85\% 
& 87.02 $\pm$ 1.39\% 
& \textbf{100.00 $\pm$ 0.00\%} \\

InversePINN~\cite{qin2024inversepinn}
& 43.92 $\pm$ 7.07\% 
& 64.18 $\pm$ 12.77\% 
& 78.12 $\pm$ 4.95\% 
& 86.84 $\pm$ 2.63\% 
& \textbf{100.00 $\pm$ 0.00\%} \\

TARTDN~\cite{cui2024tartdn}
& 60.99 $\pm$ 4.46\% 
& 60.59 $\pm$ 7.10\% 
& 69.30 $\pm$ 10.02\% 
& 90.03 $\pm$ 3.58\% 
& 98.86 $\pm$ 1.57\% \\

\bottomrule
\end{tabular}
\end{table*}

\subsection{Fault Classification Task Results}

To evaluate the classification capability of VLT under extreme label scarcity, we conduct few-shot fault diagnosis experiments on the CWRU bearing dataset under 1-shot, 2-shot, 4-shot, 8-shot, and full-data training settings, and compare its test results with FD-MVLLM~\cite{li2025fdmvllm}, LiteFormer~\cite{sun2024liteformer}, CNN-LSTM~\cite{chen2021bearing}, DP-MRTN~\cite{chen2022dualpath}, InversePINN~\cite{qin2024inversepinn}, and TARTDN~\cite{cui2024tartdn}.

\begin{figure}
    \centering
    \includegraphics[width=1\linewidth]{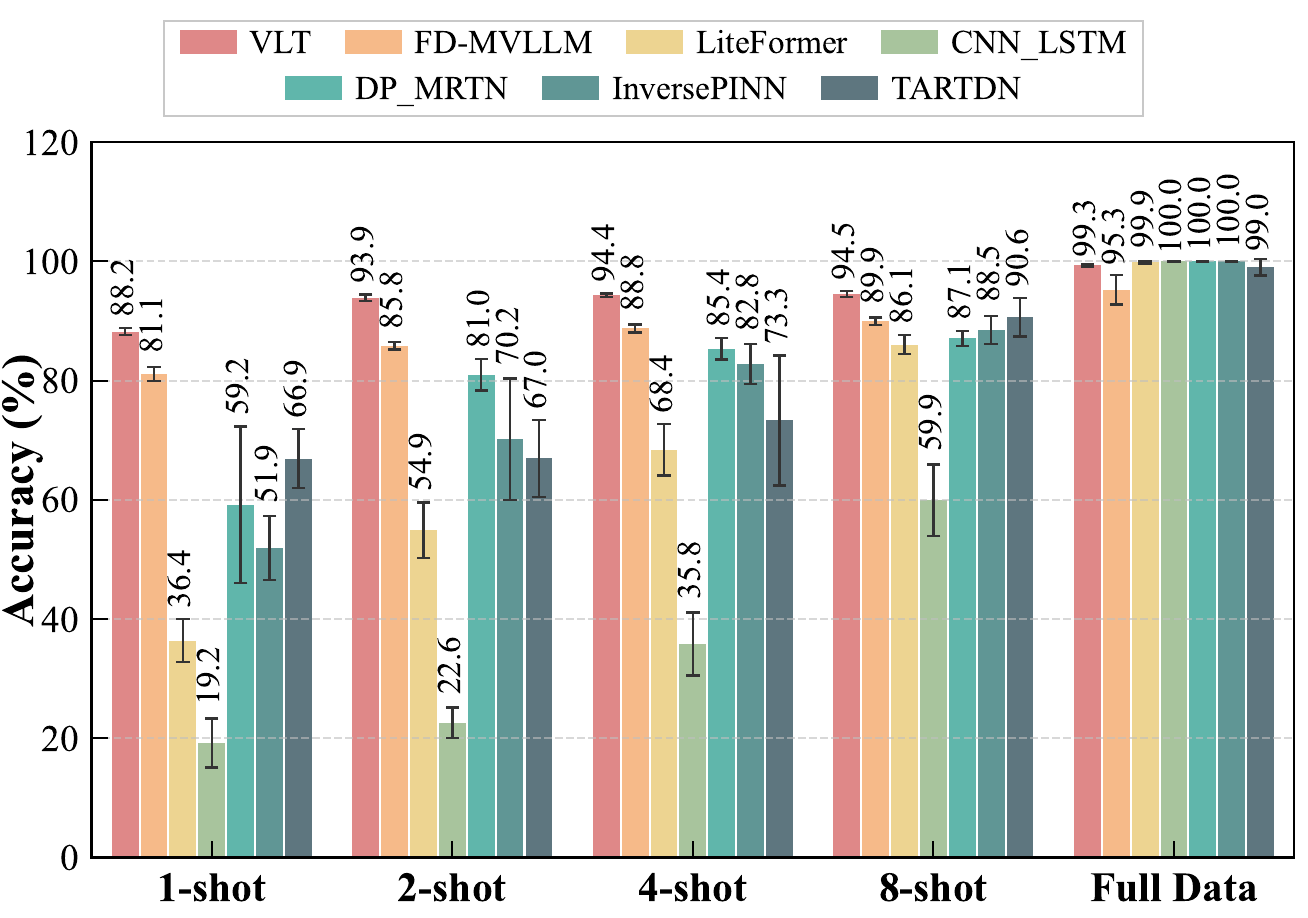}
    \caption{Fault classification task few-shot performance comparison measured by accuracy. The x-axis indicates the number of training samples ranging from 1-shot up to the full dataset.}
    \label{fig:bar_acc_single_col_labeled}
\end{figure}

As shown in Table~\ref{tab:fewshot_test_acc} and Figure~\ref{fig:bar_acc_single_col_labeled}, VLT achieves an accuracy of 88.23\%\,$\pm$\,0.59\% under the 1-shot setting. Traditional methods perform significantly worse: TARTDN achieves 66.94\%, DP\_MRTN reaches 59.17\%, and CNN\_LSTM only 19.24\%. When only one sample is available per class, models without pre-trained representations can hardly achieve effective generalization, whereas VLT benefits from the visual information and semantic priors embedded in its visual-language model backbone.

\begin{figure}
    \centering
    \includegraphics[width=1\linewidth]{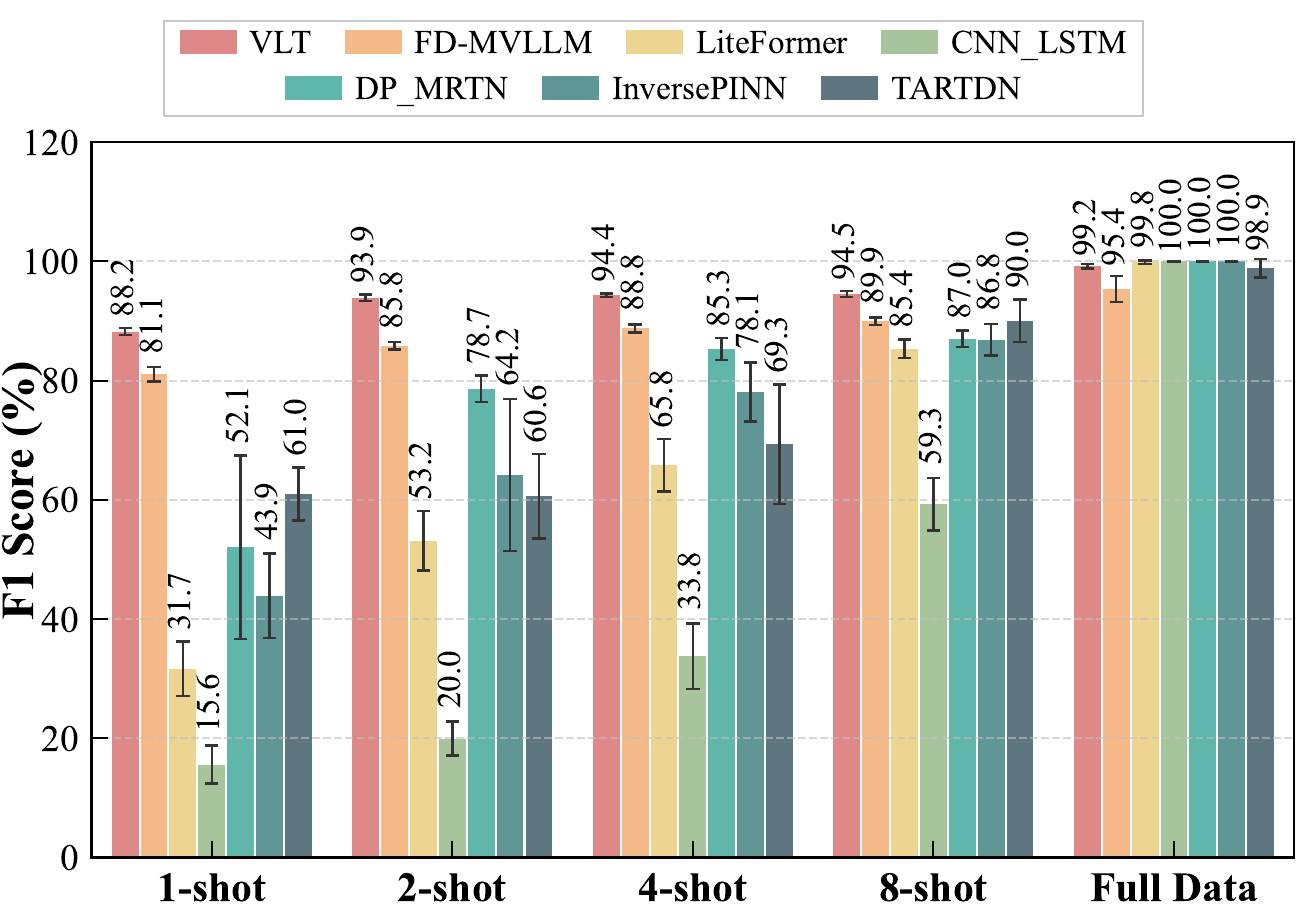}
    \caption{Fault classification task few-shot performance comparison measured by F1 score. The x-axis indicates the number of training samples ranging from 1-shot up to the full dataset.}
    \label{fig:bar_f1_single_col_labeled}
\end{figure}

As the number of samples increases, VLT improves to 93.91\% at 2-shot and reaches 94.53\% at 8-shot. Notably, with only two samples per class, VLT already outperforms the 8-shot results of TARTDN and FD-MVLLM. Meanwhile, its standard deviation remains stably within the range of 0.29\% to 0.59\%, demonstrating consistent training behavior. Under the full-data setting, CNN\_LSTM, DP\_MRTN, and InversePINN all achieve 100\% accuracy, while VLT reaches 99.33\%. Table ~\ref{tab:fewshot_test_f1} and Fig ~\ref{fig:bar_f1_single_col_labeled} further report the F1-score results, which show a trend consistent with the accuracy comparison. These results indicate that VLT not only improves overall classification correctness but also maintains balanced recognition across different fault categories.

In summary, although VLT does not achieve the optimal performance on saturated benchmark datasets, it delivers reliable and superior performance when labeled data are scarce. Since real-world industrial scenarios often suffer from a severe lack of annotated data, VLT can well meet the most critical requirements in practical industrial applications.

\begin{figure*}
    \centering
    \includegraphics[width=1\linewidth]{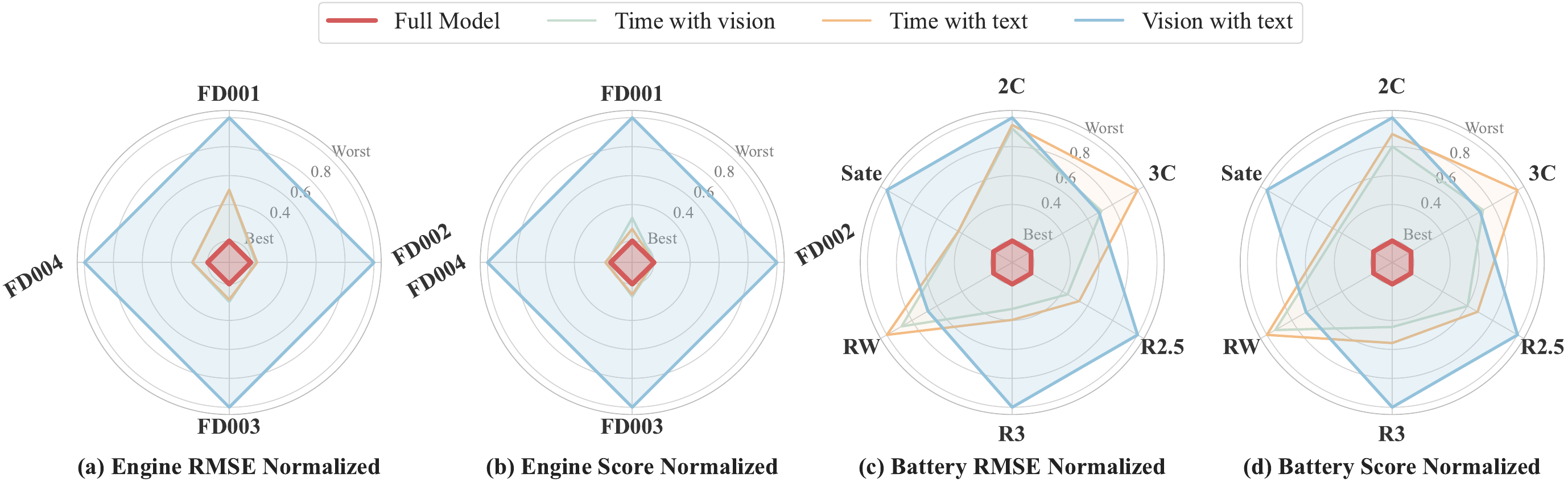}
    \caption{Radar chart of regression tasks ablation study evaluating the contribution of distinct modalities to overall predictive performance. This chart illustrates the normalized error footprint across multiple engine and battery datasets, where values closer to the center indicate lower prediction errors.}
    \label{fig:Ablation_CMPASS}
\end{figure*}

%\begin{figure*}[!t]
%    \centering
%    \includegraphics[width=1\linewidth]{new_exp_fig/ablation_combined_2x2.pdf}
%    \caption{RMSE bar charts and Score line charts obtained from ablation experiments on regression tasks. The full model is denoted in red throughout the figure.}
%    \label{fig:ablation_combined_bar}
%\end{figure*}

\begin{table*}[!t]
\centering
\caption{Ablation study for regression tasks}
\label{tab:ablation_regression}
\begin{tabular}{llcccccccc}
\toprule
\multirow{2}{*}{Category} & \multirow{2}{*}{Dataset} & \multicolumn{2}{c}{Full Model} & \multicolumn{2}{c}{Time with vision} & \multicolumn{2}{c}{Time with text} & \multicolumn{2}{c}{Vision with text} \\ 
\cmidrule(lr){3-4} \cmidrule(lr){5-6} \cmidrule(lr){7-8} \cmidrule(lr){9-10}
& & RMSE & Score & RMSE & Score & RMSE & Score & RMSE & Score \\ 
\midrule

\multirow{4}{*}{\textbf{C-MAPSS}}
& FD001   
& \textbf{11.47} & \textbf{303.70} 
& \underline{13.75} & \underline{384.70} 
& 13.76 & 345.90 
& 16.99 & 738.30 \\ 

& FD002   
& \textbf{15.48} & \textbf{1441.08} 
& \underline{16.55} & \underline{1514.63} 
& 16.86 & 1379.43 
& 42.94 & 34022.24 \\ 

& FD003   
& \textbf{11.15} & \textbf{214.40} 
& 12.04 & 271.50 
& \underline{11.92} & \underline{262.50} 
& 17.36 & 784.90 \\ 

& FD004   
& \textbf{16.59} & \textbf{1556.20} 
& 21.83 & 3585.34 
& \underline{21.70} & \underline{4245.26} 
& 43.12 & 53756.73 \\ 

\addlinespace[3pt]

\multirow{6}{*}{\textbf{Battery ($\times 10^{-1}$)}}
& 2C 
& \textbf{0.12} & \textbf{2.57} 
& \underline{0.43} & \underline{8.58} 
& 0.44 & 8.58 
& 0.46 & 8.58 \\  

& 3C 
& \textbf{0.14} & \textbf{2.93} 
& 0.43 & 9.75 
& 0.58 & 13.00 
& \underline{0.42} & \underline{9.75} \\ 

& R2.5 
& \textbf{0.07} & \textbf{1.71} 
& \underline{0.19} & \underline{8.56} 
& 0.23 & 8.56 
& 0.43 & 12.84 \\ 

& R3 
& \textbf{0.09} & \textbf{3.54} 
& \underline{0.16} & \underline{4.43} 
& 0.19 & 4.43 
& 0.43 & 13.29 \\ 

& RW 
& \textbf{0.16} & \textbf{0.83} 
& 0.47 & 5.56 
& 0.52 & 5.56 
& \underline{0.38} & \underline{4.17} \\ 

& Satellite 
& \textbf{0.06} & \textbf{2.24} 
& \underline{0.12} & \underline{5.61} 
& \underline{0.12} & \underline{5.61} 
& 0.24 & 11.22 \\ 

\bottomrule
\end{tabular}
\end{table*}

\subsection{Ablation Studies}

To evaluate the contribution of each modality branch in VLT and verify the necessity of tri-modal fusion, we perform ablation experiments for regression prediction tasks on the C-MAPSS and XJTU battery datasets, and ablation experiments for bearing fault classification on the CWRU bearing dataset. We construct three bimodal variants by sequentially removing one modality, namely Time-Vision fusion, Time-Text fusion, and Vision-Text fusion, and compare them with the full model.

\begin{table*}[b]
\centering
\caption{Ablation study for fault classification in terms of accuracy}
\label{tab:fewshot_ablation_acc}
\begin{tabular}{lccccc}
\toprule
\multirow{2}{*}{Methods} 
& \multicolumn{5}{c}{Accuracy} \\
\cmidrule(lr){2-6}
& 1-shot & 2-shot & 4-shot & 8-shot & Full Data \\
\midrule

Full Model 
& \textbf{88.23 $\pm$ 0.59\%} 
& \textbf{93.91 $\pm$ 0.54\%} 
& \textbf{94.36 $\pm$ 0.29\%} 
& \textbf{94.53 $\pm$ 0.49\%} 
& \textbf{99.33 $\pm$ 0.18\%} \\

Time with vision 
& \underline{66.23 $\pm$ 2.25\%} 
& \underline{66.84 $\pm$ 2.16\%} 
& \underline{71.95 $\pm$ 0.77\%} 
& 73.45 $\pm$ 0.15\% 
& \underline{93.60 $\pm$ 1.14\%} \\

Time with text   
& 58.82 $\pm$ 3.56\% 
& 63.95 $\pm$ 4.21\% 
& 67.88 $\pm$ 2.66\% 
& \underline{73.71 $\pm$ 1.67\%} 
& 90.03 $\pm$ 3.26\% \\

Vision with text 
& 38.05 $\pm$ 0.60\% 
& 43.40 $\pm$ 2.41\% 
& 46.37 $\pm$ 0.98\% 
& 51.61 $\pm$ 2.39\% 
& 79.85 $\pm$ 0.20\% \\

\bottomrule
\end{tabular}
\end{table*}

\begin{table*}[b]
\centering
\caption{Ablation study for fault classification in terms of F1 score}
\label{tab:fewshot_ablation_f1}
\begin{tabular}{lccccc}
\toprule
\multirow{2}{*}{Methods} 
& \multicolumn{5}{c}{F1 Score} \\
\cmidrule(lr){2-6}
& 1-shot & 2-shot & 4-shot & 8-shot & Full Data \\
\midrule

Full Model 
& \textbf{88.23 $\pm$ 0.58\%} 
& \textbf{93.91 $\pm$ 0.54\%} 
& \textbf{94.36 $\pm$ 0.29\%} 
& \textbf{94.53 $\pm$ 0.48\%} 
& \textbf{99.17 $\pm$ 0.39\%} \\

Time with vision 
& \underline{66.18 $\pm$ 2.24\%} 
& \underline{66.82 $\pm$ 2.15\%} 
& \underline{71.88 $\pm$ 0.77\%} 
& 73.42 $\pm$ 0.17\% 
& \underline{93.47 $\pm$ 1.13\%} \\

Time with text   
& 58.80 $\pm$ 3.59\% 
& 63.95 $\pm$ 4.21\% 
& 67.83 $\pm$ 2.68\% 
& \underline{73.70 $\pm$ 1.68\%} 
& 89.69 $\pm$ 3.54\% \\

Vision with text 
& 38.00 $\pm$ 0.59\% 
& 43.37 $\pm$ 2.41\% 
& 46.35 $\pm$ 1.00\% 
& 51.55 $\pm$ 2.37\% 
& 79.85 $\pm$ 0.20\% \\

\bottomrule
\end{tabular}
\end{table*}

\subsubsection{Regression Task Ablation}

Table~\ref{tab:ablation_regression}, 
Figure~\ref{fig:Ablation_CMPASS}
%, and Figure~\ref{fig:ablation_combined_bar} 
show the ablation experimental results for regression tasks. The full model achieves the lowest RMSE on all ten datasets. On FD001, the RMSE of the full model is 11.47, compared with 13.75 for the Time--Vision fusion model, 13.76 for the Time--Text fusion model, and 16.99 for the Vision--Text fusion model. The gap is even more significant on FD004: the RMSE of the full model is 16.59, while those of the three bimodal variants are 21.83, 21.70, and 43.12, respectively. For the Vision--Text fusion model with the temporal branch removed, the RMSE increases by approximately 160\%, and the Score metric surges from 1556.20 to 53756.73.

On FD002, the RMSE values of Time--Vision fusion and Time--Text fusion are 16.55 and 16.86, respectively, while the RMSE of Vision--Text fusion rises sharply to 42.94, with its Score increasing by more than 20 times.

On the R2.5\_battery dataset, the RMSE of the full model is 0.07, about one-third of that of the Time--Vision fusion model and only one-fourth of that of the Time--Text fusion model.

In summary, the two variants that retain temporal information can maintain a reasonable error range, while removing the temporal branch leads to nearly complete performance collapse. This indicates that the temporal branch serves as the core backbone of VLT, and other auxiliary modalities can also capture complementary degradation information that cannot be covered by the temporal encoder. The radar chart in Figure~\ref{fig:Ablation_CMPASS} further supports this conclusion: the contour of the full model is closest to the center, while all bimodal variants expand outward as a whole.

\subsubsection{Fault Classification Ablation}
As shown in Tables~\ref{tab:fewshot_ablation_acc} and~\ref{tab:fewshot_ablation_f1}, the complete trimodal model consistently achieves the best performance in fault classification tasks. Under the 1-shot setting, it achieves an accuracy of 88.23\% $\pm$ 0.59\% and an F1-score of 88.23\% $\pm$ 0.58\%, while removing any modality leads to an obvious performance drop. This trend remains consistent from 2-shot to 8-shot.

\begin{figure*}
    \centering
    \includegraphics[width=1\linewidth]{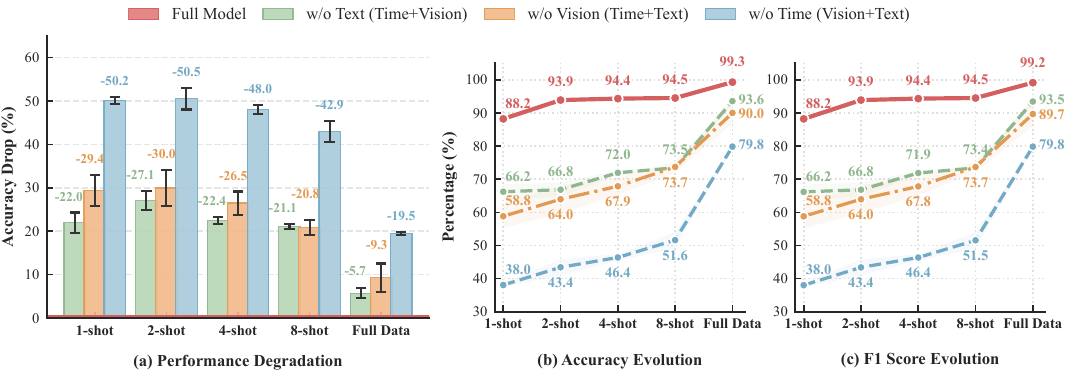}
    \caption{Ablation results on the CWRU dataset. Subfigure (a) illustrates the performance degradation caused by removing individual modalities, where the horizontal red line denotes the zero-drop baseline of the complete Full Model and the bar heights indicate the corresponding accuracy drop. Black error bars represent the variance across trials. Subfigures (b) and (c) show the absolute performance evolution of the ablated models in terms of Accuracy and F1 Score, respectively, as the training data increases from 1-shot to full data. The solid red line represents the Full Model, while the remaining curves correspond to variants with one modality removed.}
    \label{fig:ablation_cls_combined}
\end{figure*}

As shown in Fig.~\ref{fig:ablation_cls_combined}(a), removing any modality causes a noticeable degradation in classification accuracy on the CWRU dataset. The temporal modality has the largest effect. Under the 1-shot setting, removing the temporal branch leads to an accuracy drop of 50.2 percentage points, while removing the visual and textual branches causes drops of 29.4 and 22.0 percentage points, respectively. Although the degradation gradually decreases as the training data increase, it remains evident even under the Full Data setting, indicating that each modality contributes useful and non-redundant information.

Fig.~\ref{fig:ablation_cls_combined}(b) and Fig.~\ref{fig:ablation_cls_combined}(c) further show the absolute performance evolution in terms of Accuracy and F1 Score. Across all training regimes, the Full Model consistently achieves the best performance, while the ablated variants remain below it. The gap is especially large in the few-shot settings, which suggests that multimodal fusion is particularly beneficial when labeled data are limited.

In summary, the three modalities provide complementary information. Removing any one of them leads to reduced accuracy and F1-score, indicating that each modality contributes effective and non-redundant information.

\subsection{Cross-domain Generalization}
To evaluate the cross-domain generalization ability of VLT, we design two transfer tasks on the C-MAPSS dataset, namely FD001$\rightarrow$FD003 and FD003$\rightarrow$FD001. In both settings, the model is trained on the source domain and directly tested on the target domain.

As shown in Table~\ref{tab: Cross-domain performance_comparison}, VLT achieves the best overall results on both transfer tasks. For the FD001$\rightarrow$FD003 task, it achieves an RMSE of 15.733 and a Score of 442.4, significantly outperforming all baseline methods. In particular, compared with the strongest competitor PatchTST, VLT reduces the RMSE from 20.641 to 15.733 and optimizes the Score from 648.6 to 442.4, demonstrating stronger robustness under cross-domain degradation pattern shifts. In contrast, DLinear suffers severe performance degradation, with an RMSE of 63.422 and a Score of 57229, indicating that simple linear mapping is insufficient to achieve reliable cross-domain transfer.

For the FD003$\rightarrow$FD001 task, VLT still ranks first with an RMSE of 13.217 and a Score of 238.6. Its performance is slightly better than that of GPT4TS (RMSE 13.240, Score 240.8), while maintaining a more distinct advantage over other baselines.

In summary, VLT can maintain stable transfer capability even when the source and target domains are reversed. It can capture transferable degradation information and achieve more effective generalization in cross-domain scenarios.

\begin{table}% 注意这里去掉了星号，变为单栏排版
\centering
\caption{Cross-domain generalization results under different transfer tasks}
\label{tab: Cross-domain performance_comparison}
% 此时列数少，无需 \resizebox 强行缩放，保持原生字体大小更美观
\begin{tabular}{lcccc}
\toprule
\multirow{2}{*}{Methods} & \multicolumn{2}{c}{FD001 $\rightarrow$ FD003} & \multicolumn{2}{c}{FD003 $\rightarrow$ FD001} \\
\cmidrule(lr){2-3} \cmidrule(lr){4-5} % lr 参数让两条横线中间有断开，更美观
& RMSE & Score & RMSE & Score \\ 
\midrule
VLT  & \textbf{15.733} & \textbf{442.4} & \textbf{13.217} & \textbf{238.6} \\
Time-LLM~\cite{jin2024time} & 23.667 & 2287.6 & 15.328 & 347 \\
GPT4TS~\cite{zhou2023one}   & 28.290 & 6536.8 & \underline{13.240} & \underline{240.8} \\
DLinear~\cite{zeng2023transformers}  & 63.422 & 57229 & 32.043 & 2123.1 \\
PatchTST~\cite{nie2023patchtst} & \underline{20.641} & \underline{648.6} & 16.142 & 422.2 \\
TimesNet~\cite{wu2023timesnet} & 20.994 & 1597.4 & 15.019 & 421.2 \\
\bottomrule
\end{tabular}
\end{table}

\subsection{Visualization}

To investigate what VLT learns at the representation level, we present three complementary visualizations: UMAP projections of the multimodal feature space, attention fusion weight distributions, and t-SNE embeddings of few-shot classification features.

\subsubsection{Multimodal Feature Space Visualization}
\begin{figure*}
    \centering
    \includegraphics[width=1\linewidth]{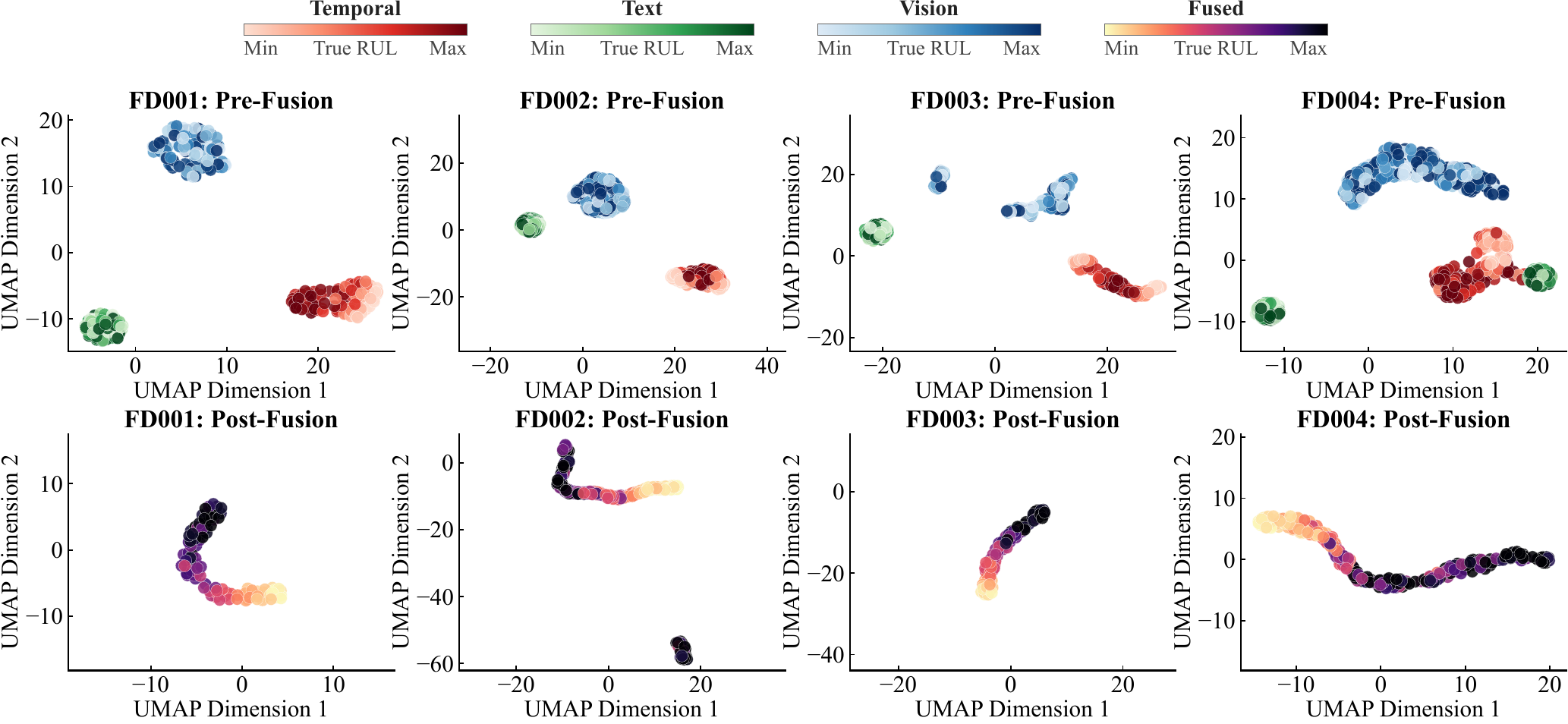}
    \vspace{-0.5em}
    \caption{UMAP visualization of multimodal feature spaces on C-MAPSS datasets (FD001--FD004). The top row shows the pre-fusion embeddings for each modality (Temporal, Text, Vision) and their concatenation, where features appear scattered and poorly structured. The bottom row presents the post-fusion feature space, revealing well-organized, continuous trajectories that reflect the progressive engine degradation process. Color gradients encode the RUL labels from minimum (blue) to maximum (red).}
    \label{fig:CMAPSS_UMAP}
\end{figure*}

Figure~\ref{fig:CMAPSS_UMAP} presents the UMAP projections of the C-MAPSS dataset before and after fusion. Before fusion, the embeddings from different modalities are scattered and heavily overlapped, and the textual and visual branches show almost no clear structure when considered separately. After fusion, the samples form smooth and continuous trajectories, and the color gradient changes monotonically with RUL, indicating that the fused space has learned to arrange samples according to degradation stages.

Figure 1 in the supplementary material shows the corresponding battery-domain visualization, and its detailed analysis is provided there.

In summary, in VLT, the original embeddings of different modalities before fusion are scattered, ambiguously structured, and severely overlapped, failing to effectively characterize the degradation patterns of equipment.After multimodal fusion, the feature space forms continuous degradation trajectories that vary monotonically with the remaining useful life (RUL), enabling the orderly arrangement of samples according to degradation stages.

\subsubsection{Attention Fusion Weight Analysis}

\begin{figure*}
    \centering
    \includegraphics[width=1\linewidth]{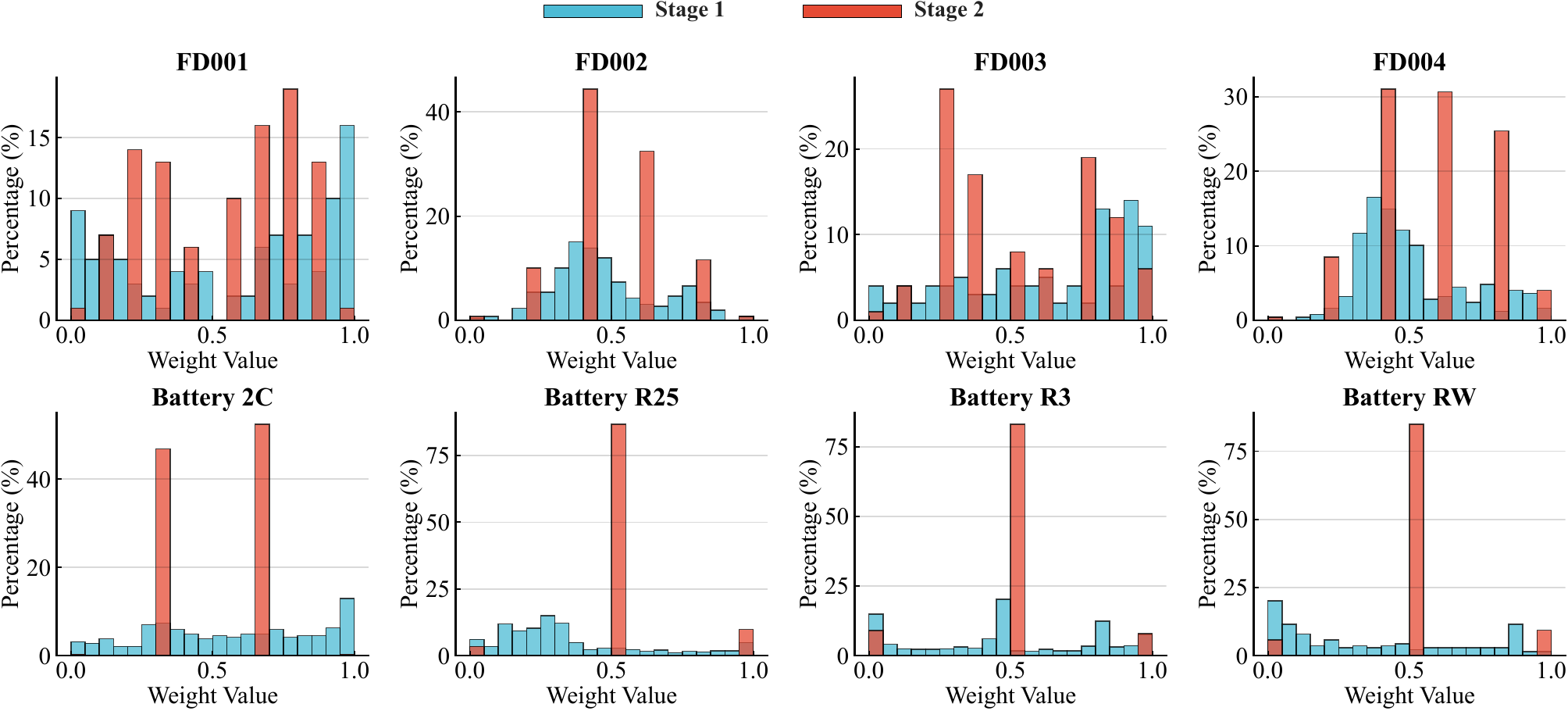}
    \vspace{-0.5em}
    \caption{Distribution of attention fusion weights across all datasets. The top row shows the weight distributions for C-MAPSS engine datasets (FD001--FD004) and one battery dataset, while the bottom row covers the remaining battery datasets. Blue and red bars represent the weights assigned to different modality fusion branches. The varying distributions across datasets demonstrate that the attention mechanism adaptively allocates modality importance based on task-specific characteristics.}
    \label{fig:weight}
\end{figure*}
\begin{figure*}
    \centering
    \includegraphics[width=1\linewidth]{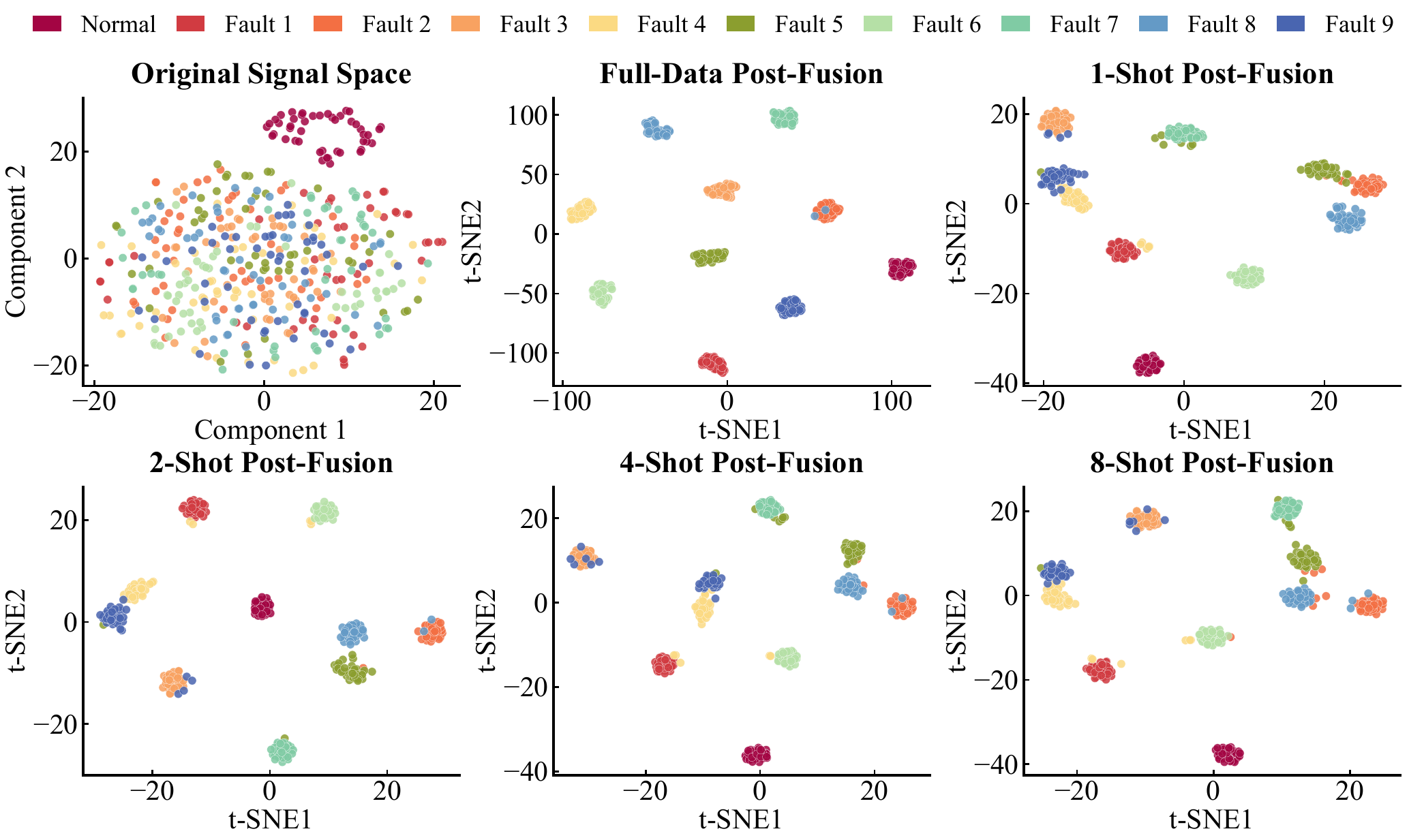}
    \vspace{-0.5em}
    \caption{t-SNE visualization of feature representations on the CWRU bearing fault dataset. Top-left: the original signal space where 10 fault categories are heavily overlapped. Top-middle and top-right: post-fusion feature spaces with full data and 1-shot, respectively. Bottom row: post-fusion feature spaces under 2-shot, 4-shot, and 8-shot settings. Each color represents a distinct fault type (Normal and Fault 1--9). The progressive improvement in cluster separation from 1-shot to full data demonstrates VLT's ability to learn discriminative multimodal representations even under extreme data scarcity.}
    \label{fig:CWRU_tSNE}
\end{figure*}

Figure~\ref{fig:weight} shows the distribution of attention-based fusion weights on different datasets under the two-stage training strategy proposed in this paper. In the first stage, only the temporal modality is trained. The weight distribution at this stage is relatively smooth and dispersed, indicating that the attention mechanism mainly performs feature reweighting within the temporal branch and has not yet formed obvious cross-modal selection ability.In the second stage, the temporal branch is frozen, and the visual and textual modalities are introduced. The dispersion of the weight distribution is significantly reduced, and clearer clustering patterns appear in local regions.

In summary, the model can selectively enhance or suppress information from different modalities according to the input features. Meanwhile, the differences in weight distribution patterns across datasets further demonstrate that the framework can dynamically adjust the contribution of each modality based on task characteristics and degradation patterns, thereby improving the adaptability of multimodal fusion.

\subsubsection{Few-shot Classification Feature Visualization}

Figure~\ref{fig:CWRU_tSNE} presents the t-SNE embeddings of fault classification features for the CWRU bearing dataset. In the original signal space, samples of the ten fault categories are heavily overlapped with no clear class boundaries. After multimodal fusion, well-defined clusters are observed under the Full Data setting. For few-shot scenarios, discernible clustering patterns already emerge under the 1-shot setting, and the clusters become increasingly compact as the number of labeled samples increases from 2-shot, 4-shot and 8-shot to Full Data, gradually converging to the clustering performance achieved under full data conditions.

Overall, these results demonstrate that the proposed VLT learns a well-structured discriminative feature space, which enables the model to correctly recognize unseen samples with extremely limited supervision. This capability is of great practical significance for real-world industrial deployment, where high-quality labeled fault data are typically scarce and difficult to obtain.

\section{Conclusion}

This paper introduced VLT, a multimodal vision-language framework for industrial time-series prediction and fault diagnosis. By jointly characterizing temporal dynamics, frequency-domain visual cues, and textual knowledge, the proposed framework constructs a unified representation space for effective cross-modal information fusion.Experiments on C-MAPSS, XJTU battery, and CWRU bearing datasets demonstrate that VLT achieves strong performance in full-sample prediction, few-shot regression, cross-domain transfer, and few-shot fault classification. Ablation and visualization results further verify that multimodal fusion improves feature structure, robustness, and generalization under data-scarce conditions.

Several limitations remain. The current framework mainly focuses on numerical prediction and classification, while natural-language diagnostic reasoning is still underexplored. Future work will extend VLT toward time-series-to-text question answering, allowing the model to answer maintenance-oriented questions from sensor sequences and provide interpretable textual explanations for industrial PHM applications.

\bibliographystyle{ieeetr}
\bibliography{reference}

\begin{IEEEbiography}[{\includegraphics[width=1in,height=2.5in,clip,keepaspectratio]{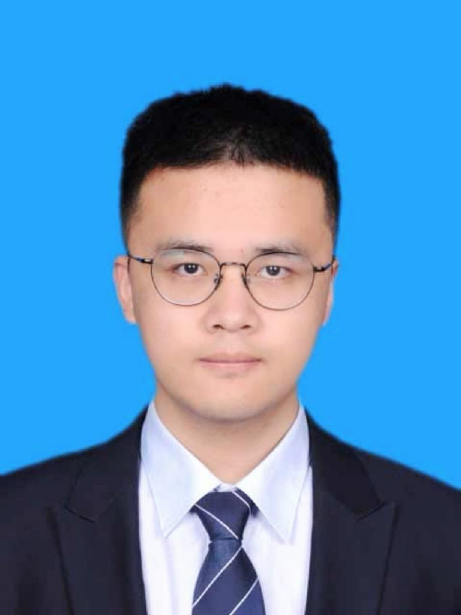}}]{Haiteng Wang} (Member, IEEE) received the Ph.D. degree in pattern recognition and intelligent systems from the School of Automation Science and Electrical Engineering, Beihang University, Beijing, China, in 2026. He is currently a Post-Doctoral Fellow with Beihang University. His current research interests include Time Series Prediction, Industrial Foundation Model, Generative Models.
\end{IEEEbiography}

\begin{IEEEbiography}[{\includegraphics[width=1in,height=2.5in,clip,keepaspectratio]{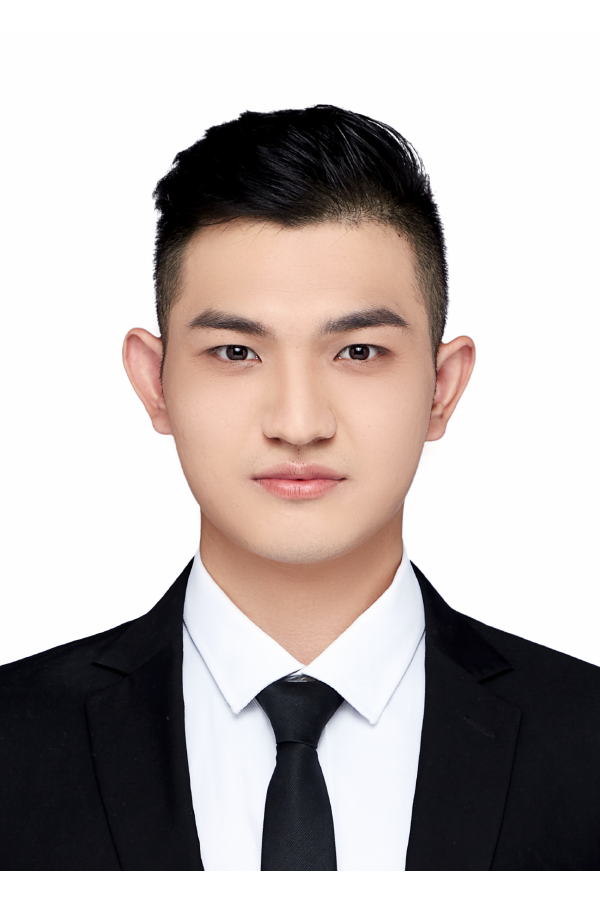}}]{Jingheng Yan} (Graduate Student Member, IEEE) received the B.Eng. Degree in automation from the School of Control Science and Engineering, Shandong University, Jinan, Shandong Province, China, in 2023. He is currently working toward the Ph.D. degree in Control Science and Engineering with the School of Automation Science and Electrical Engineering, Beihang University. His current research interests include Multi-Modal model.
\end{IEEEbiography}

\begin{IEEEbiography}[{\includegraphics[width=1in,height=2.5in,clip,keepaspectratio]{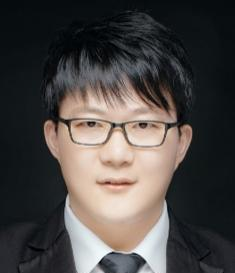}}]{Xiaokang Wang} (Member, IEEE) received the Ph.D. degree in computer system architecture from the Huazhong University of Science and Technology, Wuhan, China, in 2017. He is currently a professor with School of Computer Science and Artificial Intelligence, Zhengzhou University, China His research interests include Cyber-Physical-Social Systems, Parallel and Distributed Computing, Big Data, and Industrial Internet-of-Things.
\end{IEEEbiography}

\begin{IEEEbiography}[{\includegraphics[width=1in,height=2.5in,clip,keepaspectratio]{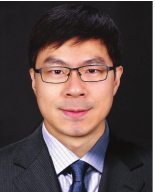}}]{Lei Ren} (Senior Member, IEEE) received the Ph.D. degree in computer science from the Institute of Software, Chinese Academy of Sciences, Beijing, China, in 2009. 
	
He is currently a Professor with the School of Automation Science and Electrical Engineering, Beihang University, Beijing, China, also with the Hangzhou International Innovation Institute, Beihang University, Hangzhou, China, and also with the State Key Laboratory of Intelligent Manufacturing System Technology, Beijing, China. His research interests include neural networks and deep learning, time series analysis, and industrial AI applications. Dr. Ren serves as an Associate Editor for the IEEE Transactions on Neural Networks and Learning Systems, IEEE/ASME Transactions on Mechatronics and other international journals.	
\end{IEEEbiography}

% ==================================================
% Supplementary Material
% ==================================================

\clearpage
\appendices

% 重新开始图表编号
\setcounter{figure}{0}
\setcounter{table}{0}

% 附录图表编号：S1、S2、S3……
\renewcommand{\thefigure}{S\arabic{figure}}
\renewcommand{\thetable}{S\arabic{table}}

% 避免 hyperref 因正文和附录编号重复而产生警告
\renewcommand{\theHfigure}{supplementary.figure.\arabic{figure}}
\renewcommand{\theHtable}{supplementary.table.\arabic{table}}

\markboth{Supplementary Material}{Supplementary Material}

\input{supplementary}

\end{document}

%% file: supplementary.tex
\markboth{Supplementary Material}{Supplementary Material}
\title{Supplementary Material for VLT}
\author{}
\maketitle
\markboth{Supplementary Material}{Supplementary Material}

\appendices

\markboth{Supplementary Material}{Supplementary Material}
\title{Supplementary Material for VLT}
\author{}
\maketitle
\markboth{Supplementary Material}{Supplementary Material}

\appendices

\section{Additional Visualization}

\begin{figure*}
    \centering
    \includegraphics[width=\linewidth]{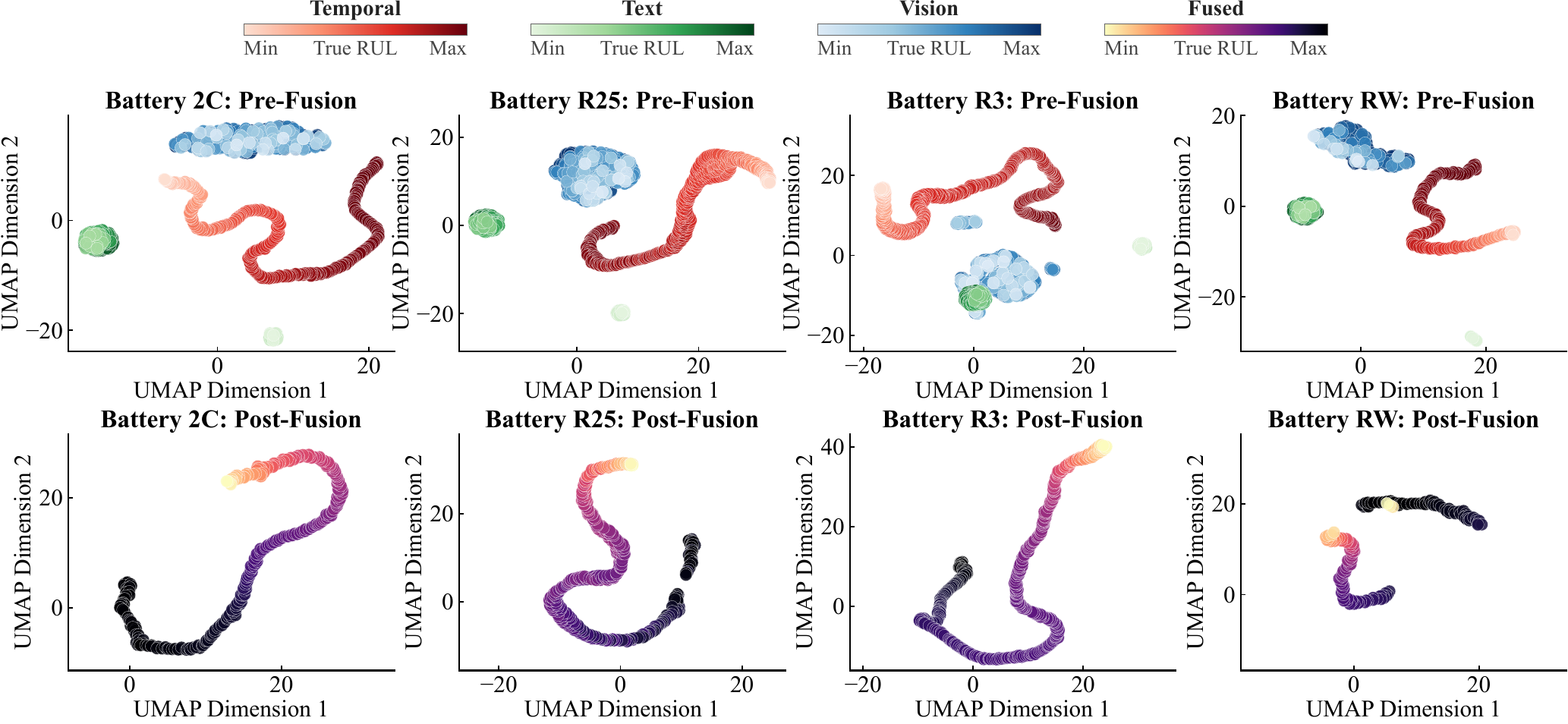}
    \vspace{-0.5em}
    \caption{UMAP visualization of multimodal feature spaces on six battery degradation datasets. The top row depicts the pre-fusion embeddings, which appear fragmented and lack clear structure. The bottom row shows the post-fusion representations, where the features form smooth, continuous curves that accurately reflect the battery capacity degradation trajectories. Color gradients indicate the state-of-health progression from early life (blue) to end-of-life (red).}
    \label{fig:Battery_UMAP}
\end{figure*}

Figure~\ref{fig:Battery_UMAP} provides the detailed battery-domain feature-space analysis. Before multimodal fusion, the embeddings of different modalities are fragmented and do not form clear degradation trajectories, indicating that individual modalities cannot fully capture the continuous capacity fading process. After fusion, the representations become more compact and are organized into smooth curves whose color gradients change consistently with the state-of-health progression. This pattern is consistent across the six battery subsets, suggesting that the proposed fusion mechanism learns a transferable degradation representation rather than relying on dataset-specific memorization.

\FloatBarrier
\section{Dataset Overview}
This supplementary material provides dataset statistics and additional visualization results that are omitted from the main paper due to space constraints.

\begin{table}
\centering
\caption{Description of the C-MAPSS dataset}
\label{tab:cmapss}
\renewcommand{\arraystretch}{1}
\begin{tabular}{lcccc}
\toprule
\textbf{Sub-dataset} & \textbf{FD001} & \textbf{FD002} & \textbf{FD003} & \textbf{FD004} \\
\midrule
Training engines     & 100 & 260 & 100 & 249 \\
Testing engines      & 100 & 259 & 100 & 248 \\
Operating conditions & 1   & 6   & 1   & 6   \\
Fault modes          & 1   & 1   & 2   & 2   \\
\bottomrule
\end{tabular}
\end{table}

\textbf{C-MAPSS Dataset:} The C-MAPSS dataset, released by NASA, is a standard benchmark in the field of aero-engine remaining useful life (RUL) prediction. The dataset comprises four subsets, FD001--FD004, which correspond to different operating conditions and fault modes. Specifically, FD001 and FD003 are defined under a single operating condition, whereas FD002 and FD004 involve six distinct operating conditions. In terms of fault modes, FD001 and FD002 contain a single fault mode, while FD003 and FD004 represent coupled degradation processes involving two fault modes. Therefore, FD002 and FD004 provide more challenging multi-condition scenarios and are useful for evaluating the robustness of PHM-VLT under complex degradation patterns.

\begin{table}
\centering
\caption{Description of the XJTU battery dataset}
\label{tab:xjtu}
\renewcommand{\arraystretch}{1}
\begin{tabular}{lcccccc}
\toprule
\textbf{Sub-dataset} & \textbf{2C} & \textbf{3C} & \textbf{R2.5} & \textbf{R3} & \textbf{RW} & \textbf{SATE} \\
\midrule
Training samples       & 2569 & 2922 & 3850 & 3983 & 1243 & 5048 \\
Testing samples        & 286  & 325  & 428  & 443  & 139  & 561  \\
Protocol category      & CR   & CR   & RD   & RD   & RW   & Sate \\
Fixed current / C-rate & 2C   & 3C   & 2.5 A & 3 A & Variable & Specific \\
Number of cells        & 8    & 15   & 8    & 8    & 8    & 8    \\
\bottomrule
\end{tabular}
\vspace{1mm}
\begin{flushleft}
\footnotesize \textit{Note:} CR = constant-rate, RD = random discharge, RW = random walking, and Sate = satellite profile.
\end{flushleft}
\end{table}

\textbf{XJTU Battery Dataset:} The XJTU battery dataset is a lithium-ion battery cycling aging dataset that records the full-life degradation trajectories of batteries under different charge-discharge strategies. In this study, six representative operating conditions, denoted as 2C, 3C, R2.5, R3, RW, and Satellite, are selected. These conditions correspond to different charge/discharge rates and complex usage scenarios, including constant-rate cycling, random discharging, and aerospace operating conditions, and are used to evaluate the model's generalization capability under multi-condition settings. The diversity of protocols makes this dataset suitable for testing whether the model can transfer degradation knowledge across different battery usage patterns.

\begin{table}
\centering
\caption{\textsc{Description of the CWRU dataset}}
\label{tab:cwru}
\begin{tabular}{cll}
\toprule
\textbf{Category} & \textbf{Parameter} & \textbf{Setting / Description} \\
\midrule
\multirow{4}{*}{Original Dataset}
 & Load conditions & 0, 1, 2, 3 hp \\
 & Typical speeds  & 1797, 1772, 1750, 1730 rpm \\
 & Sampling rates  & 12 kHz, 48 kHz \\
 & Sensor positions & Drive End, Fan End \\
\midrule
\multirow{4}{*}{10-Class Subset}
 & Health states   & \begin{tabular}[t]{@{}l@{}}Normal, Inner-race\\ Ball, Outer-race\end{tabular} \\
 & Fault sizes     & 7, 14, 21 mil \\
 & Samples per class & 467 \\
 & Train/Val/Test split & 374 / 47 / 46 \\
\bottomrule
\end{tabular}
\end{table}

\textbf{CWRU Bearing Dataset:} The CWRU bearing dataset is employed to construct the fault classification task. In this study, the drive-end vibration signals sampled at 12 kHz are selected to conduct a 10-class classification experiment. This subset includes one normal condition and nine fault conditions. The fault categories are defined by the combination of fault location and fault size, where the fault locations include the inner race, rolling element, and outer race, and the fault sizes are 7 mil, 14 mil, and 21 mil. This setting allows us to examine whether PHM-VLT can learn discriminative fault representations when only a few labeled samples are available for each class.

%% file: reference.bib
@article{wang2024sea,
  title={{SEA++}: Multi-Graph-Based Higher-Order Sensor Alignment for Multivariate Time-Series Unsupervised Domain Adaptation},
  author={Wang, Yucheng and Xu, Yuecong and Yang, Jianfei and Wu, Min and Li, Xiaoli and Xie, Lihua and Chen, Zhenghua},
  journal={IEEE Transactions on Pattern Analysis and Machine Intelligence},
  volume={46},
  number={12},
  pages={10781--10796},
  year={2024},
  publisher={IEEE},
  doi={10.1109/TPAMI.2024.3444904}
}

@article{zhu2026rul,
  title={Causality-Preserving Domain Generalization via Adaptive Fourier Mixup for {RUL} Prediction},
  author={Zhu, Yifan and Chen, Wenyu and Cheng, Zhe and Ye, Zhishen and Zhang, Fode},
  journal={IEEE Transactions on Pattern Analysis and Machine Intelligence},
  pages={1--17},
  year={2026},
  publisher={IEEE},
  note={Early Access},
  doi={10.1109/TPAMI.2026.3688520}
}

@article{ho2025graph,
  title={Graph Anomaly Detection in Time Series: A Survey},
  author={Ho, Thi Kieu Khanh and Karami, Ali and Armanfard, Narges},
  journal={IEEE Transactions on Pattern Analysis and Machine Intelligence},
  volume={47},
  number={8},
  pages={6990--7009},
  year={2025},
  publisher={IEEE},
  doi={10.1109/TPAMI.2025.3566620}
}

@article{zhang2024ssl,
  title={Self-Supervised Learning for Time Series Analysis: Taxonomy, Progress, and Prospects},
  author={Zhang, Kexin and Wen, Qingsong and Zhang, Chaoli and Cai, Rongyao and Jin, Ming and Liu, Yong and Zhang, James Y. and Liang, Yuxuan and Pang, Guansong and Song, Dongjin and Pan, Shirui},
  journal={IEEE Transactions on Pattern Analysis and Machine Intelligence},
  volume={46},
  number={10},
  pages={6775--6794},
  year={2024},
  publisher={IEEE},
  doi={10.1109/TPAMI.2024.3387317}
}

@article{zhou2023one,
  title={One fits all: Power general time series analysis by pretrained lm},
  author={Zhou, Tian and Niu, Peisong and Sun, Liang and Jin, Rong and others},
  journal={Advances in neural information processing systems},
  volume={36},
  pages={43322--43355},
  year={2023}
}

@inproceedings{zeng2023transformers,
  title={Are Transformers Effective for Time Series Forecasting?},
  author={Zeng, Ailing and Chen, Muxi and Zhang, Lei and Xu, Qiang},
  booktitle={Proceedings of the AAAI Conference on Artificial Intelligence},
  volume={37},
  number={9},
  pages={11121--11128},
  year={2023},
  doi={10.1609/aaai.v37i9.26317}
}

@article{wang2025collm,
  title={Collm: Industrial large-small model collaboration with fuzzy decision-making agent and self-reflection},
  author={Wang, Haiteng and Ren, Lei and Zhao, Tuo and Jiao, Lu},
  journal={IEEE Transactions on Fuzzy Systems},
  year={2025},
  publisher={IEEE}
}

@inproceedings{jin2024time,
  title={Time-LLM: Time Series Forecasting by Reprogramming Large Language Models},
  author={Jin, Ming and Wang, Shiyu and Ma, Lintao and Chu, Zhixuan and Zhang, James and Shi, Xiaoming and Chen, Pin-Yu and Liang, Yuxuan and Li, Yuan-fang and Pan, Shirui and others},
  booktitle={International Conference on Learning Representations},
  year={2024}
}

@article{xue2023promptcast,
  title={Promptcast: A new prompt-based learning paradigm for time series forecasting},
  author={Xue, Hao and Salim, Flora D},
  journal={IEEE Transactions on Knowledge and Data Engineering},
  volume={36},
  number={11},
  pages={6851--6864},
  year={2023},
  publisher={IEEE}
}

@inproceedings{wang2025chattime,
  title={Chattime: A unified multimodal time series foundation model bridging numerical and textual data},
  author={Wang, Chengsen and Qi, Qi and Wang, Jingyu and Sun, Haifeng and Zhuang, Zirui and Wu, Jinming and Zhang, Lei and Liao, Jianxin},
  booktitle={Proceedings of the AAAI Conference on Artificial Intelligence},
  volume={39},
  number={12},
  pages={12694--12702},
  year={2025}
}

@inproceedings{liu2024timer,
  title={Timer: generative pre-trained transformers are large time series models},
  author={Liu, Yong and Zhang, Haoran and Li, Chenyu and Huang, Xiangdong and Wang, Jianmin and Long, Mingsheng},
  booktitle={Proceedings of the 41st International Conference on Machine Learning},
  pages={32369--32399},
  year={2024}
}

@article{ansari2024chronos,
  title={Chronos: Learning the Language of Time Series},
  author={Ansari, Abdul Fatir and Stella, Lorenzo and Turkmen, Caner and Zhang, Xiyuan and Mercado, Pedro and Shen, Huibin and Shchur, Oleksandr and Rangapuram, Syama Syndar and Pineda Arango, Sebastian and Kapoor, Shubham and Zschiegner, Jasper and Maddix, Danielle C. and Mahoney, Michael W. and Torkkola, Kari and Gordon Wilson, Andrew and Bohlke-Schneider, Michael and Wang, Yuyang},
  journal={Transactions on Machine Learning Research},
  issn={2835-8856},
  year={2024},
  url={https://openreview.net/forum?id=gerNCVqqtR}
}

@inproceedings{liu2025sundial,
  title={Sundial: A Family of Highly Capable Time Series Foundation Models},
  author={Liu, Yong and Qin, Guo and Shi, Zhiyuan and Chen, Zhi and Yang, Caiyin and Huang, Xiangdong and Wang, Jianmin and Long, Mingsheng},
  booktitle={Forty-second International Conference on Machine Learning},
year={2024},
}

@inproceedings{das2024decoder,
  title={A decoder-only foundation model for time-series forecasting},
  author={Das, Abhimanyu and Kong, Weihao and Sen, Rajat and Zhou, Yichen},
  booktitle={Forty-first International Conference on Machine Learning},
year={2024},
}

@article{lan2025gem,
  title={Gem: Empowering mllm for grounded ecg understanding with time series and images},
  author={Lan, Xiang and Wu, Feng and He, Kai and Zhao, Qinghao and Hong, Shenda and Feng, Mengling},
  journal={The annual Neural Information Processing Systems conference (NIPS)},
  year={2025}
}

@article{chen2025domain,
  title={A Domain Knowledge-Guided Industrial Large Model Framework: A Case Study in Battery Health Estimation and Recycling},
  author={Chen, Bingyang and Shao, Haidong and Qin, Yao and Jin, Yang and Hu, Xinming},
  journal={IEEE Transactions on Industrial Informatics},
  year={2025},
  publisher={IEEE}
}

@article{wang2025diagllm,
  title={DiagLLM: multimodal reasoning with large language model for explainable bearing fault diagnosis},
  author={Wang, Jie and Li, Tianrui and Yang, Yan and Chen, Shiqian and Zhai, Wanming},
  journal={Science China Information Sciences},
  volume={68},
  number={6},
  pages={160103},
  year={2025},
  publisher={Springer}
}

@article{hurst2024gpt,
  title={Gpt-4o system card},
  author={Hurst, Aaron and Lerer, Adam and Goucher, Adam P and Perelman, Adam and Ramesh, Aditya and Clark, Aidan and Ostrow, AJ and Welihinda, Akila and Hayes, Alan and Radford, Alec and others},
  journal={arXiv preprint arXiv:2410.21276},
  year={2024}
}

@article{yang2025qwen3,
  title={Qwen3 technical report},
  author={Yang, An and Li, Anfeng and Yang, Baosong and Zhang, Beichen and Hui, Binyuan and Zheng, Bo and Yu, Bowen and Gao, Chang and Huang, Chengen and Lv, Chenxu and others},
  journal={arXiv preprint arXiv:2505.09388},
  year={2025}
}

@article{guo2025deepseek,
  title={Deepseek-r1 incentivizes reasoning in llms through reinforcement learning},
  author={Guo, Daya and Yang, Dejian and Zhang, Haowei and Song, Junxiao and Wang, Peiyi and Zhu, Qihao and Xu, Runxin and Zhang, Ruoyu and Ma, Shirong and Bi, Xiao and others},
  journal={Nature},
  volume={645},
  number={8081},
  pages={633--638},
  year={2025},
  publisher={Nature Publishing Group UK London}
}

@article{zhang2023data,
  title={A data-model interactive remaining useful life prediction approach of lithium-ion batteries based on PF-BiGRU-TSAM},
  author={Zhang, Jiusi and Huang, Congsheng and Chow, Mo-Yuen and Li, Xiang and Tian, Jilun and Luo, Hao and Yin, Shen},
  journal={IEEE Transactions on Industrial Informatics},
  volume={20},
  number={2},
  pages={1144--1154},
  year={2023},
  publisher={IEEE}
}

@article{jin2023adaptive,
  title={An adaptive and dynamical neural network for machine remaining useful life prediction},
  author={Jin, Ruibing and Zhou, Duo and Wu, Min and Li, Xiaoli and Chen, Zhenghua},
  journal={IEEE Transactions on Industrial Informatics},
  volume={20},
  number={2},
  pages={1093--1102},
  year={2023},
  publisher={IEEE}
}

@ARTICLE{ren2024mts,
  author={Ren, Lei and Wang, Haiteng and Laili, Yuanjun},
  journal={IEEE Transactions on Cybernetics}, 
  title={Diff-MTS: Temporal-Augmented Conditional Diffusion-Based AIGC for Industrial Time Series Toward the Large Model Era}, 
  year={2024},
  volume={54},
  number={12},
  pages={7187-7197},
  doi={10.1109/TCYB.2024.3462500}}

@article{zhou2021novel,
  title={A novel soft sensor modeling approach based on difference-LSTM for complex industrial process},
  author={Zhou, Jiayi and Wang, Xiaoli and Yang, Chunhua and Xiong, Wei},
  journal={IEEE Transactions on Industrial Informatics},
  volume={18},
  number={5},
  pages={2955--2964},
  year={2021},
  publisher={IEEE}
}

@article{wang2025meta,
  title={MetaIndux-TS: Frequency-Aware AIGC Foundation Model for Industrial Time Series},
  author={Wang, Haiteng and Ren, Lei and Li, Yikang},
  journal={IEEE Transactions on Neural Networks and Learning Systems},
  pages={1--13},
  year={2021},
  publisher={IEEE}
}

@ARTICLE{ren2023dynamic,
  author={Ren, Lei and Wang, Haiteng and Huang, Gao},
  journal={IEEE Transactions on Neural Networks and Learning Systems}, 
  title={DLformer: A Dynamic Length Transformer-Based Network for Efficient Feature Representation in Remaining Useful Life Prediction}, 
  year={2024},
  volume={35},
  number={5},
  pages={5942-5952},
  doi={10.1109/TNNLS.2023.3257038}}

@inproceedings{yu2023harnessing,
  title={Harnessing LLMs for Temporal Data-A Study on Explainable Financial Time Series Forecasting},
  author={Yu, Xinli and Chen, Zheng and Lu, Yanbin},
  booktitle={Proceedings of the 2023 Conference on Empirical Methods in Natural Language Processing: Industry Track},
  pages={739--753},
  year={2023}
}

@article{tao2025llm,
  title={LLM-based framework for bearing fault diagnosis},
  author={Tao, Laifa and Liu, Haifei and Ning, Guoao and Cao, Wenyan and Huang, Bohao and Lu, Chen},
  journal={Mechanical Systems and Signal Processing},
  volume={224},
  pages={112127},
  year={2025},
  publisher={Elsevier}
}

@article{sun2025fine,
  title={Fine-tuning enables state of health estimation for lithium-ion batteries via a time series foundation model},
  author={Sun, Wenjie and Wu, Chengke and Xie, Chengde and Wang, Xikang and Guo, Yuanjun and Tang, Yongbing and Zhang, Yanhui and Li, Kang and Du, Guanhao and Yang, Zhile and others},
  journal={Energy},
  volume={318},
  pages={134177},
  year={2025},
  publisher={Elsevier}
}

@article{zhao2025predictive,
  title={Predictive pretrained transformer (PPT) for real-time battery health diagnostics},
  author={Zhao, Jingyuan and Wang, Zhenghong and Wu, Yuyan and Burke, Andrew F},
  journal={Applied Energy},
  volume={377},
  pages={124746},
  year={2025},
  publisher={Elsevier}
}

@article{li2025fdmvllm,
  title={FD-MVLLM: Fault Diagnosis Based on Multimodal Vibration Data and Large Language Model for Bearing System},
  author={Li, Dayang and Pang, Zhibo and Chen, Yongxu and Yang, Kun and Shao, Jinyan and Luo, Yichen and Zeng, Yanghang and He, Chen and Gao, Yutong},
  journal={Mechanical Systems and Signal Processing},
  volume={239},
  pages={113226},
  year={2025},
  doi={10.1016/j.ymssp.2025.113226},
  publisher={Elsevier}
}

@article{sun2024liteformer,
  title={LiteFormer: A Lightweight and Efficient Transformer for Rotating Machine Fault Diagnosis},
  author={Sun, Wenjun and Yan, Ruqiang and Jin, Ruibing and Xu, Jiawen and Yang, Yuan and Chen, Zhenghua},
  journal={IEEE Transactions on Reliability},
  volume={73},
  number={2},
  pages={1258--1269},
  year={2024},
  doi={10.1109/TR.2023.3322860},
  publisher={IEEE}
}

@article{chen2021bearing,
  title={Bearing Fault Diagnosis Base on Multi-Scale CNN and LSTM Model},
  author={Chen, Xiaohan and Zhang, Beike and Gao, Dong},
  journal={Journal of Intelligent Manufacturing},
  volume={32},
  number={4},
  pages={971--987},
  year={2021},
  doi={10.1007/s10845-020-01600-2},
  publisher={Springer}
}

@article{chen2022dualpath,
  title={Dual-Path Mixed-Domain Residual Threshold Networks for Bearing Fault Diagnosis},
  author={Chen, Yongyi and Zhang, Dan and Zhang, Hui and Wang, Qing-Guo},
  journal={IEEE Transactions on Industrial Electronics},
  volume={69},
  number={12},
  pages={13462--13472},
  year={2022},
  doi={10.1109/TIE.2022.3144572},
  publisher={IEEE}
}

@article{qin2024inversepinn,
  title={Inverse Physics-Informed Neural Networks for Digital Twin-Based Bearing Fault Diagnosis Under Imbalanced Samples},
  author={Qin, Yi and Liu, Hongyu and Wang, Yi and Mao, Yongfang},
  journal={Knowledge-Based Systems},
  volume={292},
  pages={111641},
  year={2024},
  doi={10.1016/j.knosys.2024.111641},
  publisher={Elsevier}
}

@article{cui2024tartdn,
  title={Triplet Attention-Enhanced Residual Tree-Inspired Decision Network: A Hierarchical Fault Diagnosis Model for Unbalanced Bearing Datasets},
  author={Cui, Lingli and Dong, Zhilin and Xu, Hai and Zhao, Dezun},
  journal={Advanced Engineering Informatics},
  volume={59},
  pages={102322},
  year={2024},
  doi={10.1016/j.aei.2023.102322},
  publisher={Elsevier}
}

@article{ouyang2024combined,
  title={Combined meta-learning with CNN-LSTM algorithms for state-of-health estimation of lithium-ion battery},
  author={Ouyang, Tiancheng and Su, Yingying and Wang, Chengchao and Jin, Song},
  journal={IEEE Transactions on Power Electronics},
  volume={39},
  number={8},
  pages={10106--10117},
  year={2024},
  publisher={IEEE}
}

@article{yuan2025lithium,
  title={A lithium-ion battery state of health estimation method utilizing convolutional neural networks and bidirectional long short-term memory with attention mechanisms for collaborative defense against false data injection cyber-attacks},
  author={Yuan, Tianqing and Gao, Feng and Bai, Jing and Sun, Hao},
  journal={Journal of Power Sources},
  volume={631},
  pages={236193},
  year={2025},
  publisher={Elsevier}
}

@article{jin2022bi,
  title={Bi-LSTM-based two-stream network for machine remaining useful life prediction},
  author={Jin, Ruibing and Chen, Zhenghua and Wu, Keyu and Wu, Min and Li, Xiaoli and Yan, Ruqiang},
  journal={IEEE Transactions on Instrumentation and Measurement},
  volume={71},
  pages={1--10},
  year={2022},
  publisher={IEEE}
}

@inproceedings{zheng2017lstm,
  title={Long Short-Term Memory Network for Remaining Useful Life Estimation},
  author={Zheng, Shuai and Ristovski, Kosta and Farahat, Ahmed and Gupta, Chetan},
  booktitle={2017 IEEE International Conference on Prognostics and Health Management (ICPHM)},
  pages={88--95},
  year={2017},
  doi={10.1109/ICPHM.2017.7998311}
}

@inproceedings{wen2018cnnfd,
  title={A New Convolutional Neural Network-Based Data-Driven Fault Diagnosis Method},
  author={Wen, Long and Li, Xinyu and Gao, Liang and Zhang, Yuyan},
  journal={IEEE Transactions on Industrial Electronics},
  volume={65},
  number={7},
  pages={5990--5998},
  year={2018},
  doi={10.1109/TIE.2017.2774777}
}

@article{jiang2019multiscalecnn,
  title={Multiscale Convolutional Neural Networks for Fault Diagnosis of Wind Turbine Gearbox},
  author={Jiang, Guoqian and He, Haibo and Yan, Jun and Xie, Ping},
  journal={IEEE Transactions on Industrial Electronics},
  volume={66},
  number={4},
  pages={3196--3207},
  year={2019},
  doi={10.1109/TIE.2018.2844805}
}

@article{chen2020tcnn,
  title={Intelligent Fault Diagnosis for Rotary Machinery Using Transferable Convolutional Neural Network},
  author={Chen, Zhuyun and Gryllias, Konstantinos and Li, Weihua},
  journal={IEEE Transactions on Industrial Informatics},
  volume={16},
  number={1},
  pages={339--349},
  year={2020},
  doi={10.1109/TII.2019.2917233}
}

@article{ardakani2024narx,
  title={NARX Transformer: A Dynamic Model for Leveraging Multicycle Data in Long-Term Battery State of Health Estimation},
  author={Ardakani, Amirhossein Heydarian and Abdollahian, Seyed Ali and Abdollahi, Farzaneh},
  journal={IEEE Transactions on Instrumentation and Measurement},
  volume={73},
  pages={2530908},
  year={2024},
  doi={10.1109/TIM.2024.3460947}
}

@inproceedings{jia2024gpt4mts,
  title={GPT4MTS: Prompt-Based Large Language Model for Multimodal Time-Series Forecasting},
  author={Jia, Furong and Wang, Kevin and Zheng, Yixiang and Cao, Defu and Liu, Yan},
  booktitle={Proceedings of the AAAI Conference on Artificial Intelligence},
  volume={38},
  number={21},
  pages={23343--23351},
  year={2024},
  doi={10.1609/aaai.v38i21.30383}
}

@inproceedings{nie2023patchtst,
  title={A Time Series is Worth 64 Words: Long-term Forecasting With Transformers},
  author={Nie, Yuqi and Nguyen, Nam H. and Sinthong, Phanwadee and Kalagnanam, Jayant},
  booktitle={International Conference on Learning Representations},
  year={2023},
  url={https://openreview.net/forum?id=Jbdc0vTOcol}
}

@inproceedings{wu2023timesnet,
  title={TimesNet: Temporal 2D-Variation Modeling for General Time Series Analysis},
  author={Wu, Haixu and Hu, Tengge and Liu, Yong and Zhou, Hang and Wang, Jianmin and Long, Mingsheng},
  booktitle={International Conference on Learning Representations},
  year={2023},
  url={https://openreview.net/forum?id=ju_Uqw384Oq}
}
